\documentclass[11pt]{article}
\usepackage{graphicx}
\usepackage{url}
\usepackage{amsmath}
\usepackage{amssymb}
\usepackage{amsfonts}
\usepackage{xcolor}
\usepackage{array}
\usepackage{colortbl}
\usepackage{float}
\usepackage{subcaption}
\usepackage{caption}
\usepackage{booktabs}
\usepackage{tabularx}
\usepackage{geometry}
\usepackage{multirow}
\usepackage{hyperref}
\usepackage{microtype}
\usepackage{authblk}
\usepackage{algorithm}
\usepackage{algorithmic}
\usepackage{tikz}
\usetikzlibrary{arrows.meta, positioning, fit, backgrounds}
\usepackage[numbers,sort&compress]{natbib}

\title{Repair Lot Skyline: A Weighted Constraint Satisfaction Approach to Pavement Repair Optimization from Geospatial Hazard Density}

\author[1]{Takato Yasuno\thanks{Corresponding author and lead contributor. Email: \texttt{tkt-yasuno@yachiyo-eng.co.jp}}}
\author[1]{Keita Kobayashi}
\author[1]{Ryuta Sakaguchi}
\author[1]{Takuya Okamoto}
\affil[1]{Digital Business Promotion Department, Yachiyo Engineering Co., Ltd.}

\date{\today}

\begin{document}

\maketitle

\begin{abstract}

Pavement agencies must translate a spatially distributed distress inventory into a bounded, actionable repair-lot plan: accident-critical defects (potholes) must always be addressed, lower-risk defects (cracks) should be included only when their benefit justifies the repair cost, and historical patch locations signal re-degradation risk without themselves triggering repair. We formalize this as a \emph{Repair Lot Skyline} problem: a Weighted Constraint Satisfaction Problem (WCSP) defined over chainage (distance along the road) rather than over time, so that it requires only a single-epoch distress survey and makes no claim about future deterioration.

\textbf{Approach:} Distress observations are clustered along chainage into candidate lots; a hard constraint forces any lot containing a pothole into the repair plan; a soft constraint admits a crack-only lot only when a calibrated benefit-to-cost threshold is met; patch observations are attached as a non-binding historical recurrence-risk feature; and every lot is padded, using real chainage distance rather than image count, to a minimum serviceable length. We further construct a cumulative \emph{skyline} frontier that orders soft candidates by benefit-per-metre efficiency, exposing the full budget-versus-coverage trade-off rather than a single operating point.

\textbf{Case Study:} We apply this method to the publicly available PaveDistress image collection (S315 highway, China), a single-epoch, spatially sparse survey whose 4,922 deduplicated unit images span a nominal chainage range of 1,541 km but cover only 48 km of \emph{actually surveyed} road across four disjoint sections. Kilopost-level geospatial hazard density is visualized using a fixed z-score scaling of pothole/crack/patch counts, consistent with scaling-robustness evidence from a related deterioration-modeling study that is otherwise out of scope here.

\textbf{Key Findings:} The WCSP identifies 143 candidate hazard clusters (61 hard, 82 soft), of which 106 are merged into a final repair plan totaling 1,997.4 m---83.7\% of the 2,385.9 m that would be required if every soft candidate were included regardless of cost. This plan covers 100\% of observed potholes (138/138) and 91.6\% of observed cracks (404/441), capturing 93.6\% (542/579) of the total hazard benefit available in the full candidate set. The skyline frontier shows pronounced diminishing returns beyond this point: the remaining 37 excluded soft candidates would add only 6.8\% additional benefit for a 19.4\% increase in repair length. We further formalize the minimum-lot-length $L_{\min}$ and historical-context radius $\kappa$ as a joint, four-objective hyperparameter search over this WCSP; on the same case study, the recommended configuration ($L_{\min} = 14.7$ m, $\kappa = 10$ m) reduces repair-crew mobilizations by 7.1\% relative to an untuned default, at the cost of a 12.9\% larger budget and a 1.1-percentage-point lower crack coverage.

\textbf{Contributions:} (1) A spatial WCSP formulation of repair-lot selection with explicit hard/soft/historical semantics, distinct from and complementary to deterioration-prediction modeling. (2) A real-distance lot-padding procedure that guarantees a minimum serviceable repair length under arbitrarily sparse spatial sampling. (3) A skyline frontier construction that reframes repair-budget selection as a decision-support surface rather than a single heuristic threshold. (4) A joint, multi-objective hyperparameter search over the minimum-lot-length and historical-context radius, including the finding that a mobilization-count objective is necessary to prevent the search from collapsing to a degenerate, over-fragmented optimum. (5) A fixed, geospatial hazard density representation (Definitions 2--3) that aggregates per-image distress labels into a kilopost-level, z-score-scaled map along chainage for decision support, without forecasting future deterioration.

\end{abstract}

\section{Introduction}
\label{sec:introduction}

\subsection{Background and Motivation}

Road agencies increasingly rely on image-based distress surveys---collected by inspection vehicles, mobile phones, or dashboard cameras---to assess pavement condition over long highway networks~\citep{Cao2020}. These surveys typically yield a large, spatially indexed collection of defect observations (potholes, cracks, patches) rather than a continuous condition curve. Converting this raw distress inventory into an actionable \emph{repair plan}---a finite set of contiguous road segments, or \emph{lots}, that a maintenance crew will actually mobilize to repair under a bounded budget---is a distinct and under-formalized decision problem from the deterioration-\emph{prediction} problem addressed by classical pavement management systems~\citep{Golabi1982,Shahin2005} and Markov-based infrastructure deterioration models~\citep{FrangopolLiu2007}.

Three practical considerations make repair-lot planning a constraint satisfaction problem rather than a pure prediction or ranking problem. First, some defect types are safety-critical: a pothole is a documented cause of vehicle damage and accidents, so any lot containing one should be repaired essentially unconditionally, independent of cost. Second, other defect types are lower-risk and discretionary: a crack-only location is worth repairing only when the achievable benefit (number of distress observations addressed) exceeds the cost of mobilizing a repair crew over the affected length. Third, some information is historical context rather than an actionable signal: a prior patch at a location indicates the site is prone to re-degradation, which is useful for prioritization and monitoring but should not by itself trigger a new repair action. A repair-planning method that treats all three considerations identically---for example, by ranking all defects by a single severity score and thresholding---cannot represent this qualitative distinction between \emph{mandatory}, \emph{discretionary}, and \emph{contextual} information.

\subsection{Problem Statement and Challenges}

We ask: \textbf{given a single-epoch, spatially indexed pavement distress survey, how should a finite set of repair lots be selected so that safety-critical defects are always addressed, discretionary defects are included only when cost-justified, and the resulting plan is transparent to a maintenance planner who must operate under a flexible, not fixed, budget?} We term this the \emph{Repair Lot Skyline} problem and formulate it as a Weighted Constraint Satisfaction Problem (WCSP) defined over chainage---the one-dimensional distance coordinate along a road, also called a kilopost---rather than over time.

This formulation surfaces three challenges beyond a naive distress-count threshold.

\textbf{Challenge 1: Heterogeneous constraint semantics.} Pothole, crack, and patch observations must be treated with three qualitatively different decision rules (hard, soft, and historical-context, respectively) within a single optimization, not three independent rankings that must later be reconciled manually.

\textbf{Challenge 2: Sparse and irregular spatial sampling.} A real distress survey does not sample chainage uniformly; gaps between observations vary by orders of magnitude, and a repair lot constructed naively from a short run of consecutive images can be shorter than the minimum length a repair crew can mobilize for. A valid lot-construction procedure must guarantee a minimum \emph{real-distance} length regardless of local sampling density, and must do so without assuming the survey covers the full nominal length of the road.

\textbf{Challenge 3: No single ``optimal'' budget.} The cost/benefit threshold that separates included from excluded discretionary (crack) lots is a modeling choice, not a ground truth; a planner facing a different budget year may reasonably want a smaller or larger repair plan. A method that reports only one plan, with no visibility into the trade-off curve it was selected from, does not support this kind of what-if reasoning.

\subsection{Proposed Approach}

We formalize chainage-adjacent pothole/crack observations as hazard clusters (Section~\ref{sec:methodology}), apply a hard constraint that unconditionally includes any cluster containing a pothole, apply a soft constraint that includes a crack-only cluster only when a calibrated benefit-to-cost threshold is met, and attach nearby patch observations to each lot as a non-binding historical recurrence-risk annotation. Every candidate lot is padded, using real chainage distance rather than image count, to a minimum serviceable repair length. Finally, we construct a cumulative \emph{skyline} frontier that orders discretionary candidates by benefit-per-metre efficiency, so that the full budget-versus-coverage trade-off---not just the single operating point selected by the default threshold---is available to a planner.

We deliberately scope this contribution to \emph{decision support from a single-epoch spatial survey}, not deterioration prediction. The PaveDistress dataset used as our case study (Section~\ref{sec:experiments}) provides no repeated inspection of the same physical location over time; it is a one-time spatial inventory. A companion line of work by the same authors separately studies Markov deterioration hazard modeling from (pseudo-)temporal bridge and pavement inspection transitions~\citep{Yasuno2026Scaling}; that modeling problem, its assumptions, and its results are unrelated to and out of scope for the spatial decision-support method presented here.

\subsection{Key Contributions}

\begin{enumerate}
    \item \textbf{A formal WCSP specification of repair-lot selection} with explicit hard (safety-critical), soft (discretionary, cost/benefit-gated), and historical-context (non-binding) constraint semantics, distinguishing this decision-support formulation from deterioration-prediction modeling.
    \item \textbf{A real-distance minimum-lot-length procedure} that guarantees every selected repair lot reaches a minimum serviceable length by padding in physical distance rather than image count, so that sparse or irregular spatial sampling cannot silently produce an unrealistically short repair lot.
    \item \textbf{A skyline frontier construction} that exposes the full cumulative budget-versus-benefit trade-off of discretionary (crack) inclusion, reframing repair-plan selection as an inspectable decision surface rather than a single opaque threshold.
    \item \textbf{A joint, multi-objective hyperparameter search} over the minimum-lot-length $L_{\min}$ and historical-context radius $\kappa$, formalizing both as tunable parameters subject to a four-objective Pareto search (budget, mobilization count, crack coverage, patch specificity); we show that a mobilization-count objective is necessary to prevent the search from collapsing to a degenerate, over-fragmented optimum (Section~\ref{sec:hyperparam-opt}).
    \item \textbf{A fixed, geospatial hazard density representation} (Definitions 2--3) that aggregates per-image pothole/crack/patch labels into a kilopost-level, z-score-scaled hazard map along chainage, giving planners a descriptive, decision-support view of \emph{current} distress density rather than a forecast of future deterioration (Section~\ref{sec:notation}).
\end{enumerate}

\subsection{Paper Organization}

Section~\ref{sec:related} reviews related work on constraint satisfaction, skyline queries, and pavement management optimization. Section~\ref{sec:methodology} presents the formal WCSP model: hazard clustering, minimum-lot-length expansion, historical-context attachment, the hard/soft decision rule, lot merging, the skyline frontier, and the $(L_{\min}, \kappa)$ hyperparameter search. Section~\ref{sec:experiments} describes the PaveDistress case-study dataset, the hyperparameter grid-search setup, and the implementation. Section~\ref{sec:results} reports the resulting hazard clusters, repair plan, skyline frontier, and hyperparameter search outcome. Section~\ref{sec:discussion} interprets the findings, discusses limitations, and gives practical guidance. Section~\ref{sec:conclusion} concludes.

\section{Related Work}
\label{sec:related}

\subsection{Constraint Satisfaction and Weighted CSP}

Classical constraint satisfaction problems (CSPs) require every constraint to be satisfied exactly~\citep{RossiVanBeekWalsh2006}. Many real-world decision problems, however, mix constraints that must always hold with preferences that should be satisfied when affordable. The Valued CSP / Weighted CSP (WCSP) framework formalizes this distinction by associating each constraint with a cost (zero for satisfaction, a finite penalty for violation, and infinite for a hard constraint that can never be violated), and seeks an assignment minimizing total cost~\citep{Schiex1995,Bistarelli1997}. Exact WCSP solvers typically rely on branch-and-bound with soft local-consistency enforcement~\citep{LarrosaMorancho2005}, which scales well on the type of loosely coupled, chain-structured candidate lot problems that arise from spatial clustering, but general-purpose exact solvers are not required for our setting: because our chainage-ordered candidate lots do not interact once clustered (Section~\ref{sec:clustering}), a single left-to-right pass evaluating each cluster's hard/soft status independently is exact for our model. We adopt the WCSP vocabulary---hard constraints, soft constraints with a cost/benefit trade-off, and non-binding auxiliary information---because it precisely captures the heterogeneous semantics required by pothole, crack, and patch observations (Section~\ref{sec:introduction}), while keeping the solution procedure itself simple and transparent to a non-specialist planner.

A substantial line of work develops general-purpose exact and metaheuristic WCSP solvers that we deliberately do not require but that establish the broader algorithmic vocabulary this paper borrows. \citet{NonobeIbaraki1998} formulate the CSP itself as a weighted optimization problem and solve it with a tabu-search general problem solver, an early demonstration that treating hard constraints as infinitely weighted soft constraints yields a single, uniform search procedure---the same unification principle underlying Eq.~\ref{eq:decision}'s hard/soft rule. Subsequent work established increasingly strong local-consistency properties for pruning the WCSP search space without losing exactness: \citet{LarrosaSchiex2003} compare arc-, directional-arc-, and full-directional-arc consistency and show that the strongest property is typically also the most effective in practice; \citet{CooperSchiex2004} give a general arc-consistency theory for soft constraints that subsumes several such properties; and \citet{DeGivryEtAl2005} introduce existential arc consistency as a cheaper approximation that recovers most of full arc consistency's pruning power. These results are representative of a mature, general-purpose WCSP solving literature; our contribution is not a new solver but a problem-specific observation that none of this machinery is needed once the hazard units are clustered along chainage, because the resulting candidate lots are not jointly constrained with one another.

\subsection{Skyline Queries and Pareto Frontiers}

The skyline operator from the database literature returns the set of records not dominated by any other record across multiple attributes, without requiring the user to specify a single scalar weighting of those attributes in advance~\citep{Borzsonyi2001}. This is closely related to the Pareto frontier in multi-objective optimization, which characterizes the full set of non-dominated trade-off points between competing objectives rather than a single optimal point~\citep{Miettinen1999}. We adopt this idea to address Challenge 3 of Section~\ref{sec:introduction}: rather than reporting only the repair plan selected by a single cost/benefit threshold, we construct the cumulative frontier of all possible discretionary-inclusion decisions, ordered by benefit-per-metre efficiency, so that a planner can inspect how much additional coverage each additional unit of budget would buy, instead of being limited to a single operating point.

The dominance test underlying both the skyline operator and the Pareto frontier predates either literature: \citet{KungLuccioPreparata1975} give the first efficient algorithm for finding the maxima of a set of vectors under componentwise dominance, the combinatorial core later rediscovered as the database skyline query. Within databases, subsequent work focused on computing the skyline efficiently over large relations without materializing and sorting the entire dominance graph: \citet{KossmannRamsakRost2002} propose an online, nearest-neighbor-search-based algorithm that returns skyline points progressively; \citet{PapadiasEtAl2003} give a branch-and-bound algorithm that is simultaneously I/O-optimal and progressive; and \citet{ChomickiEtAl2003} show that a single presorting step by a monotone scoring function makes a block-nested-loops skyline computation both simpler and faster in practice. Our skyline frontier (Definition 9--10) needs none of this machinery, since our candidate set is small and already chainage-sorted, but we adopt the same non-dominance semantics these algorithms compute. On the multi-objective optimization side, \citet{DebEtAl2002}'s NSGA-II is the dominant reference algorithm for approximating a Pareto frontier via evolutionary search when the objective space cannot be enumerated exhaustively, and \citet{Ehrgott2005} gives the standard treatment of Pareto dominance, efficiency, and scalarization for continuous and combinatorial multi-objective problems alike. Our hyperparameter search (Section~\ref{sec:hyperparam-opt}) instead evaluates a small, finite grid exhaustively (Section~\ref{sec:hyperparam-experiments}), so an evolutionary approximation such as NSGA-II is unnecessary, but we use exactly the Pareto-dominance test and a weight-free ideal-point selection rule in the tradition of this literature (Definition 12).

\subsection{Pavement and Infrastructure Maintenance Optimization}

Pavement management systems have long used Markov-chain condition models combined with linear or dynamic programming to select network-level maintenance and rehabilitation (M\&R) actions under a budget constraint, most notably the Arizona statewide pavement management system~\citep{Golabi1982} and its successors~\citep{Shahin2005}. More recent work formulates pavement preservation scheduling as a multi-objective optimization balancing cost, condition, and user-delay objectives~\citep{ChenZhou2019}, and surveys of infrastructure maintenance optimization emphasize life-cycle cost and condition-based triggering rules~\citep{FrangopolLiu2007}. These approaches operate at the level of a maintenance action selected per road segment or per bridge component over a planning horizon, typically driven by a fitted deterioration or performance curve. Our problem differs in scope and input: we do not assume access to a deterioration curve or a multi-year planning horizon, and we operate at the finer spatial granularity of individual distress observations along chainage, which must first be clustered into candidate repair lots before any selection rule can be applied (Section~\ref{sec:clustering}).

Several further studies formulate pavement M\&R scheduling explicitly as a constrained or multi-objective optimization problem over a multi-year horizon. \citet{OuyangMadanat2004} derive exact and approximate solutions for scheduling rehabilitation activities across multiple pavement facilities under a shared budget, exploiting problem structure to avoid full dynamic-programming enumeration. \citet{ChootinanEtAl2006} embed a stochastic pavement-deterioration simulation inside a genetic algorithm to produce multi-year maintenance programs that are robust to deterioration uncertainty, rather than optimized only against its expected trajectory. \citet{WuFlintsch2009} extend network-level preservation optimization to multiple, potentially conflicting objectives under budget variability using chance constraints, and \citet{LiEtAl2022} apply a multi-objective differential-evolution algorithm to jointly optimize cost, condition, and sustainability (emissions) objectives for a pavement network. Every one of these methods, like the two foundational systems above, requires a calibrated deterioration or performance-prediction model and plans over multiple future budget years; our Repair Lot Skyline WCSP instead requires only the current, single-epoch distress survey and produces an immediately actionable spatial repair plan, making it complementary to rather than a replacement for this multi-year M\&R programming literature (Section~\ref{sec:discussion}).

\subsection{Road-Surface Damage Detection Using Visual Inspection}

A substantial body of work addresses the perception problem of detecting and classifying pavement distress (cracks, potholes, patches) from images using classical image processing and, increasingly, deep learning~\citep{Cao2020}. Representative examples include convolutional neural networks trained directly on crack image patches~\citep{ZhangEtAl2016} and smartphone-camera-based road damage detectors trained on large, multi-class annotated datasets for deployment on ordinary municipal vehicles rather than specialized line-scan survey equipment~\citep{MaedaEtAl2018}. This literature treats each image independently as a classification or segmentation target and does not address how detected distress should be aggregated along chainage into a bounded set of repair actions. Our work is complementary and downstream: we take an existing, per-image category label as given (the PaveDistress dataset provides this directly, Section~\ref{sec:experiments}) and address the repair-lot aggregation and selection problem that follows detection.

\subsection{Relation to Deterioration-Prediction Modeling}

A separate line of work by the same authors studies Markov deterioration hazard modeling for infrastructure assets, including covariate scaling robustness for maximum-likelihood estimation on bridge inspection transition data~\citep{Yasuno2026Scaling}. That work requires paired before/after inspection observations of the same physical location and estimates a hazard rate governing future state transitions; it is a temporal prediction problem. This problem class is well established beyond our own prior work: \citet{MadanatMishalaniIbrahim1995} give an early and widely used method for estimating discrete-state infrastructure transition probabilities directly from condition-rating data, the statistical foundation underlying most subsequent Markovian pavement and bridge deterioration models, and \citet{FrangopolKallenVanNoortwijk2004} survey probabilistic life-cycle performance models for deteriorating structures more broadly, including both Markov-chain and stochastic-process formulations. The present paper requires no such paired observations: it operates on a single-epoch spatial distress survey and produces a repair plan, not a forecast. We note this distinction explicitly because the two problems are easily conflated when both are applied to pavement data, but they answer different questions (``what will happen'' versus ``what should be repaired now, given what is observed'') and make different assumptions about the underlying data (repeated temporal observations versus a single spatial snapshot).

\section{Methodology}
\label{sec:methodology}

This section presents the geospatial hazard representation, the hazard-clustering and minimum-lot-length procedures, the WCSP hard/soft/historical decision rule, and the skyline frontier construction, in formal notation with an explicit algorithm. Implementation notes and parameter-calibration details supporting this section are collected in Appendix~\ref{sec:appendix-methodology}. Figure~\ref{fig:overview-tikz} previews the full pipeline before the formal definitions below introduce it step by step.

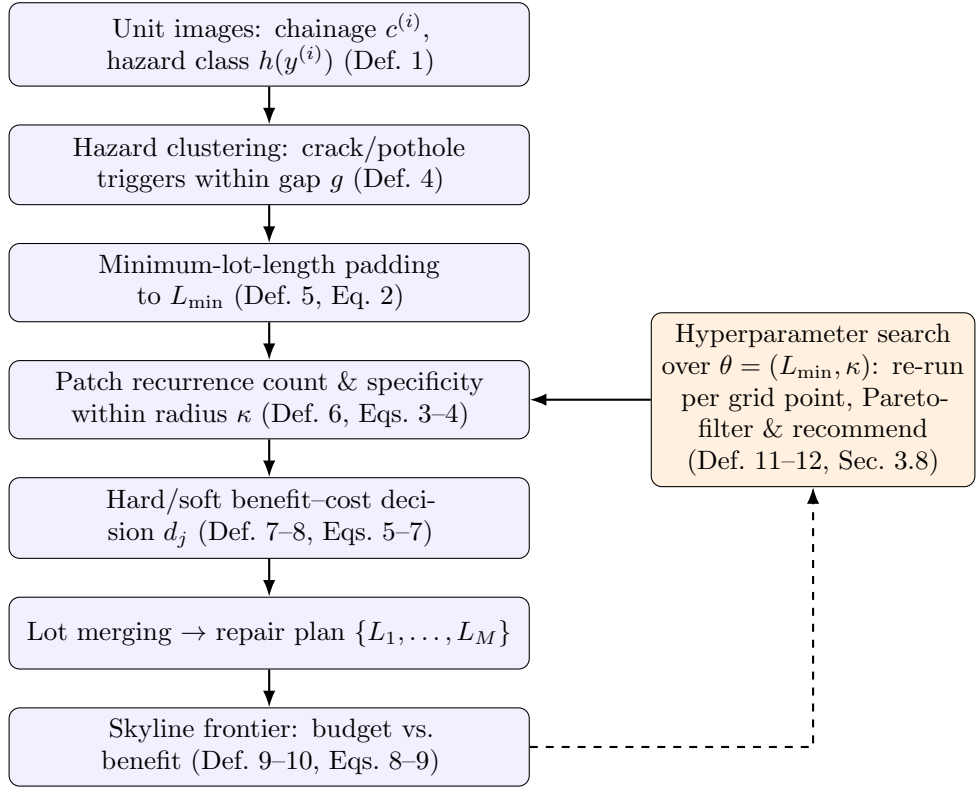
\begin{figure}[h]
\centering
\begin{tikzpicture}[
  node distance=0.5cm and 1.6cm,
  every node/.style={font=\small},
  box/.style={draw, rounded corners, align=center, fill=blue!6, text width=6.6cm, minimum height=0.95cm, inner sep=4pt},
  hbox/.style={draw, rounded corners, align=center, fill=orange!12, text width=4.0cm, minimum height=2.0cm, inner sep=4pt},
  arrow/.style={-{Latex[length=2.2mm]}, thick}
]
\node[box] (input) {Unit images: chainage $c^{(i)}$, hazard class $h(y^{(i)})$ (Def.~1)};
\node[box, below=of input] (cluster) {Hazard clustering: crack/pothole triggers within gap $g$ (Def.~4)};
\node[box, below=of cluster] (pad) {Minimum-lot-length padding to $L_{\min}$ (Def.~5, Eq.~\ref{eq:minlot})};
\node[box, below=of pad] (patch) {Patch recurrence count \& specificity within radius $\kappa$ (Def.~6, Eqs.~\ref{eq:patch-recurrence}--\ref{eq:patch-specificity})};
\node[box, below=of patch] (decide) {Hard/soft benefit--cost decision $d_j$ (Def.~7--8, Eqs.~\ref{eq:benefit-cost-generalized}--\ref{eq:decision})};
\node[box, below=of decide] (merge) {Lot merging $\rightarrow$ repair plan $\{L_1,\ldots,L_M\}$};
\node[box, below=of merge] (skyline) {Skyline frontier: budget vs.\ benefit (Def.~9--10, Eqs.~\ref{eq:skyline-baseline}--\ref{eq:skyline-step})};

\node[hbox, right=of patch] (hyper) {Hyperparameter search over $\theta = (L_{\min}, \kappa)$: re-run per grid point, Pareto-filter \& recommend (Def.~11--12, Sec.~\ref{sec:hyperparam-opt})};

\draw[arrow] (input) -- (cluster);
\draw[arrow] (cluster) -- (pad);
\draw[arrow] (pad) -- (patch);
\draw[arrow] (patch) -- (decide);
\draw[arrow] (decide) -- (merge);
\draw[arrow] (merge) -- (skyline);

\draw[arrow] (hyper.west) -- (patch.east);
\draw[arrow, dashed] (skyline.east) -| (hyper.south);
\end{tikzpicture}
\caption{Overview of the Repair-Lot Skyline WCSP (Algorithm~\ref{alg:repair-lot-skyline}): unit images are clustered, padded to a minimum serviceable length, annotated with historical patch context, and decided hard/soft/excluded; included lots are merged into the final repair plan and accumulated into the skyline frontier. The dashed loop at right indicates that Section~\ref{sec:hyperparam-opt}'s hyperparameter search re-runs this entire pipeline once per $(L_{\min}, \kappa)$ grid point.}
\label{fig:overview-tikz}
\end{figure}

\subsection{Problem Setup and Notation}
\label{sec:notation}

Let $i = 1, \ldots, n$ index unique distress-survey \emph{unit images} after deduplication (Appendix~\ref{sec:appendix-dedup}), each with a chainage (distance-along-road) coordinate $c^{(i)} \in \mathbb{R}_{\geq 0}$, expressed in metres as $c^{(i)}_m = 1000\, c^{(i)}_{\text{km}}$, and a category label $y^{(i)} \in \{\text{Background}, \text{Other}, \text{Crack}, \text{Patch}, \text{Pothole}\}$. Each unit image represents a fixed real-world footprint of width $w = 3.9$~m (across the lane) by pitch $p = 2.1$~m (along chainage), consistent with the acquisition geometry of the case-study dataset (Section~\ref{sec:experiments}).

\textbf{Definition 1 (Hazard class).} The hazard class $h: y \mapsto \{\texttt{none}, \texttt{crack}, \texttt{patch}, \texttt{pothole}\}$ maps Background and Other to \texttt{none} (no actionable distress), Crack to \texttt{crack}, Patch to \texttt{patch}, and Pothole to \texttt{pothole}. Other is mapped to \texttt{none} rather than discarded, to maximize label coverage of the survey (Appendix~\ref{sec:appendix-dedup}).

\textbf{Definition 2 (Kilopost hazard bins).} For a fixed bin width $b$ (default $b = 1$~km), define bin index $k(c) = \lfloor c / b \rfloor$ and, for each bin $k$ spanning all unit images in that bin, the count vector $N_k = (N_k^{\text{pothole}}, N_k^{\text{crack}}, N_k^{\text{patch}}, N_k^{\text{none}})$.

\textbf{Definition 3 (Fixed z-score hazard score).} For each hazard count column $N^{\tau}$, $\tau \in \{\texttt{pothole}, \texttt{crack}, \texttt{patch}\}$, let $\bar{N}^\tau$ and $s^\tau$ denote the mean and population standard deviation of $N^\tau$ over all bins of the surveyed network. The fixed z-score hazard score is
\begin{equation}
z^\tau_k = \frac{N_k^\tau - \bar{N}^\tau}{s^\tau}, \qquad
\mathrm{HazardScore}_k = z^{\text{pothole}}_k + z^{\text{crack}}_k + z^{\text{patch}}_k.
\label{eq:zscore-hazard}
\end{equation}
Although this paper makes no claim about future deterioration (Section~\ref{sec:limitations}), $\mathrm{HazardScore}_k$ is computed from the single-epoch survey to assess the \emph{current} geospatial density of pavement distress hazard along chainage, not to forecast how that density will evolve over time. The scaling method is fixed to z-score for all reported visualizations; no alternative scaling is explored in this paper, consistent with scaling-robustness evidence reported in a related, out-of-scope deterioration-modeling study~\citep{Yasuno2026Scaling} (Appendix~\ref{sec:appendix-scaling-equivalence}). Because the hazard map is a descriptive representation over one fixed road network rather than a train/test predictive split, fitting the scaler over all bins raises no evaluation-leakage concern.

\subsection{Hazard Clustering}
\label{sec:clustering}

\hspace*{1em}\textbf{Definition 4 (Trigger units and clusters).} Let $U_{\text{trigger}} = \{i : h(y^{(i)}) \in \{\texttt{crack}, \texttt{pothole}\}\}$, sorted by chainage. Patch and none-class units are not cluster triggers: patch is historical context attached later (Definition~6), and none-class units carry no distress to repair. For a fixed gap tolerance $g$ (default $g = 11$~m, calibrated from the empirical inter-observation chainage gap distribution; Appendix~\ref{sec:appendix-gap}), two chainage-consecutive trigger units $i, i{+}1 \in U_{\text{trigger}}$ belong to the same cluster iff $c^{(i+1)}_m - c^{(i)}_m \leq g$. This partitions $U_{\text{trigger}}$ into clusters $\{C_1, \ldots, C_J\}$, each with span $[\mathrm{start}(C_j), \mathrm{end}(C_j)]$, pothole count $P_j = |\{i \in C_j : h(y^{(i)}) = \texttt{pothole}\}|$, crack count $R_j = |\{i \in C_j : h(y^{(i)}) = \texttt{crack}\}|$, and hard-trigger indicator $\mathrm{Hard}_j = \mathbb{1}[P_j > 0]$.

\subsection{Minimum-Lot-Length Expansion}
\label{sec:minlot}

\hspace*{1em}\textbf{Definition 5 (Minimum serviceable lot length).} Let $L_{\min} = p \cdot u_{\min}$, where $u_{\min} = 5$ is the minimum number of pitch units a repair crew can mobilize for ($L_{\min} = 2.1 \times 5 = 10.5$~m). For a cluster $C_j$ with real length $\ell_j = \mathrm{end}(C_j) - \mathrm{start}(C_j)$, if $\ell_j < L_{\min}$, the span is symmetrically padded by $\delta_j = (L_{\min} - \ell_j)/2$ on each side:
\begin{equation}
\mathrm{start}'(C_j) = \mathrm{start}(C_j) - \delta_j, \qquad \mathrm{end}'(C_j) = \mathrm{end}(C_j) + \delta_j,
\label{eq:minlot}
\end{equation}
subject to clipping at the surveyed route boundary $[\rho_{\min}, \rho_{\max}]$; any padding deficit caused by clipping on one side is pushed entirely to the other side so that $\mathrm{end}'(C_j) - \mathrm{start}'(C_j) = L_{\min}$ whenever the surveyed route is long enough to accommodate it. Padding uses real chainage distance rather than image count, so a lot constructed from very sparsely sampled images is not silently shorter than $L_{\min}$.

\subsection{Historical Recurrence-Risk Attachment}
\label{sec:patch-history}

\hspace*{1em}\textbf{Definition 6 (Patch recurrence count and specificity).} For a context radius $\kappa$ (default $\kappa = 50$~m) and padded lot span $[\mathrm{start}'(C_j), \mathrm{end}'(C_j)]$, the patch recurrence count is
\begin{equation}
\mathrm{PatchCount}_j(\kappa) = \big|\{ i : h(y^{(i)}) = \texttt{patch},\ c^{(i)}_m \in [\mathrm{start}'(C_j) - \kappa,\ \mathrm{end}'(C_j) + \kappa] \}\big|.
\label{eq:patch-recurrence}
\end{equation}
In the baseline model of this paper, $\mathrm{PatchCount}_j(\kappa)$ is attached to every candidate lot as a non-binding annotation and never appears in the inclusion decision (Section~\ref{sec:decision}), consistent with its role as historical, re-degradation-risk context rather than an actionable repair trigger. Section~\ref{sec:hyperparam-opt} relaxes this: $\kappa$ is treated as a tunable hyperparameter and $\mathrm{PatchCount}_j(\kappa)$ is allowed a small, bounded influence on the soft-constraint benefit (Eq.~\ref{eq:benefit-cost-generalized}), without ever letting it trigger or block inclusion on its own. To assess whether a chosen $\kappa$ is counting patches that are genuinely nearby or merely numerous-but-distant, we additionally define, for a fixed, tight reference radius $\kappa_0$ (default $\kappa_0 = 15$~m),
\begin{equation}
\mathrm{Specificity}_j(\kappa) = \min\!\left(1,\ \frac{\mathrm{PatchCount}_j(\kappa_0)}{\mathrm{PatchCount}_j(\kappa)}\right), \qquad \mathrm{Specificity}_j(\kappa) := 1 \text{ if } \mathrm{PatchCount}_j(\kappa) = 0.
\label{eq:patch-specificity}
\end{equation}
By construction $\mathrm{Specificity}_j(\kappa_0) = 1$ and $\mathrm{Specificity}_j(\kappa)$ is non-increasing as $\kappa$ grows beyond $\kappa_0$: widening $\kappa$ can only ever dilute, never improve, the fraction of counted patches that are truly close to the lot, which is what makes $\kappa$ a genuine precision-versus-recall hyperparameter rather than one for which ``wider is always better.''

\subsection{WCSP Hard/Soft Decision Rule}
\label{sec:decision}

\hspace*{1em}\textbf{Definition 7 (Benefit and cost).} For soft-crack weight $w_{\text{soft}} > 0$, recurrence-risk weight $w_{\text{rec}} \geq 0$, and cost-per-metre weight $w_{\text{cost}} > 0$ (defaults $w_{\text{soft}} = 1$, $w_{\text{rec}} = 0.2$, $w_{\text{cost}} = 0.12$; calibration in Appendix~\ref{sec:appendix-cost}), define
\begin{equation}
\mathrm{Benefit}_j = w_{\text{soft}} \cdot R_j + w_{\text{rec}} \cdot \mathrm{PatchCount}_j(\kappa), \qquad \mathrm{Cost}_j = w_{\text{cost}} \cdot \big(\mathrm{end}'(C_j) - \mathrm{start}'(C_j)\big).
\label{eq:benefit-cost-generalized}
\end{equation}
Setting $w_{\text{rec}} = 0$ recovers the baseline model $\mathrm{Benefit}_j = w_{\text{soft}} \cdot R_j$ (Eq.~\ref{eq:benefit-cost}) used throughout Section~\ref{sec:results}'s primary case study; $w_{\text{rec}} \ll w_{\text{soft}}$ by design (an order of magnitude smaller), so a chronically re-patched stretch can only \emph{nudge} a borderline crack-only cluster toward inclusion, never dominate the crack-driven decision or substitute for it.
\begin{equation}
\mathrm{Benefit}_j = w_{\text{soft}} \cdot R_j, \qquad \mathrm{Cost}_j = w_{\text{cost}} \cdot \big(\mathrm{end}'(C_j) - \mathrm{start}'(C_j)\big).
\label{eq:benefit-cost}
\end{equation}

\textbf{Definition 8 (Hard/soft inclusion decision).} The WCSP decision $d_j \in \{\texttt{include}, \texttt{exclude}\}$ for lot $j$ is
\begin{equation}
d_j =
\begin{cases}
\texttt{include} & \text{if } \mathrm{Hard}_j = 1 \quad \text{(pothole hard constraint)} \\
\texttt{include} & \text{if } \mathrm{Hard}_j = 0 \text{ and } \mathrm{Benefit}_j \geq \mathrm{Cost}_j \quad \text{(crack soft constraint satisfied)} \\
\texttt{exclude} & \text{otherwise.}
\end{cases}
\label{eq:decision}
\end{equation}
Because $\mathrm{Hard}_j$ is an unconditional trigger with effectively infinite cost of exclusion, Eq.~\ref{eq:decision} is a two-level WCSP: a hard constraint with infinite violation cost (pothole presence), and a soft constraint with a finite, calibrated violation cost that is traded off against an explicit benefit (crack density per unit repair length).

\subsection{Lot Merging}

Included lots ($d_j = \texttt{include}$) are merged, in chainage order, whenever two lots overlap or touch ($\mathrm{start}'(C_{j+1}) \leq \mathrm{end}'(C_j)$), producing the final repair plan $\{L_1, \ldots, L_M\}$ with $M \leq J$. Each merged lot inherits the union of its source clusters' pothole, crack, and patch counts, and is labeled \texttt{hard} if any source cluster was hard, else \texttt{soft}.

\subsection{Skyline Frontier}
\label{sec:skyline-def}

\hspace*{1em}\textbf{Definition 9 (Benefit-per-metre efficiency).} For each soft (non-hard) cluster, define efficiency $e_j = \mathrm{Benefit}_j / \ell'_j$, where $\ell'_j$ is the padded lot length.

\textbf{Definition 10 (Cumulative skyline frontier).} Let $H$ be the set of hard clusters and $S$ the set of soft clusters sorted by $e_j$ in decreasing order. The frontier is the sequence of points $(\Lambda_t, B_t)_{t=0}^{|S|}$ with mandatory baseline
\begin{equation}
\Lambda_0 = \sum_{j \in H} \ell'_j, \qquad B_0 = \sum_{j \in H} (P_j + R_j),
\label{eq:skyline-baseline}
\end{equation}
and, for $t = 1, \ldots, |S|$, letting $j_t$ be the $t$-th cluster in sorted order,
\begin{equation}
\Lambda_t = \Lambda_{t-1} + \ell'_{j_t}, \qquad B_t = B_{t-1} + R_{j_t}.
\label{eq:skyline-step}
\end{equation}
Each point $t$ is marked \emph{selected} iff $d_{j_t} = \texttt{include}$ under Eq.~\ref{eq:decision}. The frontier $(\Lambda_t, B_t)$ is, by construction, non-decreasing in both coordinates and gives the full achievable trade-off between cumulative repair length (budget) and cumulative hazard benefit (coverage), of which the WCSP's default threshold selects one particular, but not the only inspectable, operating point.

\begin{algorithm}[H]
\caption{Repair Lot Skyline WCSP}
\label{alg:repair-lot-skyline}
\begin{algorithmic}[1]
\STATE \textbf{Input}: unit images with chainage and hazard class; parameters $g, L_{\min}, \kappa, w_{\text{soft}}, w_{\text{rec}}, w_{\text{cost}}$
\STATE \textbf{Output}: candidate lots with decisions, merged repair plan, skyline frontier
\STATE Build hazard clusters $\{C_j\}$ from crack/pothole trigger units by chainage gap $g$ (Definition 4)
\FOR{each cluster $C_j$}
    \STATE Pad to minimum length $L_{\min}$ using real chainage distance, clipped at route bounds (Eq.~\ref{eq:minlot})
    \STATE Attach patch recurrence count and specificity within context radius $\kappa$ (Eqs.~\ref{eq:patch-recurrence}--\ref{eq:patch-specificity})
    \STATE Compute benefit and cost (Eq.~\ref{eq:benefit-cost-generalized}); decide $d_j$ (Eq.~\ref{eq:decision})
\ENDFOR
\STATE Merge all \texttt{include} lots that overlap or touch into the final repair plan $\{L_1, \ldots, L_M\}$
\STATE Sort soft clusters by efficiency $e_j$ descending; accumulate the skyline frontier from the hard baseline (Eqs.~\ref{eq:skyline-baseline}--\ref{eq:skyline-step})
\end{algorithmic}
\end{algorithm}

\paragraph{Quick reference for Algorithm~\ref{alg:repair-lot-skyline}.} For convenience, the clustering rule and the six equations the algorithm invokes are restated here compactly, so they can be checked without paging back to Sections~\ref{sec:clustering}--\ref{sec:decision}:
\begin{itemize}
\setlength{\itemsep}{1pt}
\item \textbf{Def.~4 (clusters):} consecutive crack/pothole trigger units $i, i{+}1$ join the same cluster $C_j$ iff $c^{(i+1)}_m - c^{(i)}_m \leq g$.
\item \textbf{Eq.~\ref{eq:minlot} (padding):} if $\ell_j < L_{\min}$, pad symmetrically by $\delta_j = (L_{\min} - \ell_j)/2$ per side (clipped and rebalanced at route bounds).
\item \textbf{Eq.~\ref{eq:patch-recurrence} (patch count):} $\mathrm{PatchCount}_j(\kappa) = $ number of patch units within $\kappa$ metres of the padded span $[\mathrm{start}'(C_j), \mathrm{end}'(C_j)]$.
\item \textbf{Eq.~\ref{eq:patch-specificity} (specificity):} $\mathrm{Specificity}_j(\kappa) = \min(1,\ \mathrm{PatchCount}_j(\kappa_0) / \mathrm{PatchCount}_j(\kappa))$, with reference radius $\kappa_0 < \kappa$.
\item \textbf{Eq.~\ref{eq:benefit-cost-generalized} (benefit/cost):} $\mathrm{Benefit}_j = w_{\text{soft}} R_j + w_{\text{rec}} \mathrm{PatchCount}_j(\kappa)$; $\mathrm{Cost}_j = w_{\text{cost}} \cdot (\mathrm{end}'(C_j) - \mathrm{start}'(C_j))$; setting $w_{\text{rec}} = 0$ recovers the baseline Eq.~\ref{eq:benefit-cost}, $\mathrm{Benefit}_j = w_{\text{soft}} R_j$.
\item \textbf{Eq.~\ref{eq:decision} (decision):} $d_j = \texttt{include}$ if $\mathrm{Hard}_j = 1$ (pothole present) or $\mathrm{Benefit}_j \geq \mathrm{Cost}_j$; else $d_j = \texttt{exclude}$.
\end{itemize}

\subsection{Hyperparameter Optimization over $(L_{\min}, \kappa)$}
\label{sec:hyperparam-opt}

Algorithm~\ref{alg:repair-lot-skyline} is parameterized by, among others, the minimum serviceable lot length $L_{\min}$ (Definition 5) and the historical-context radius $\kappa$ (Definition 6). Rather than treat these as fixed constants chosen once by inspection, we treat $\theta = (L_{\min}, \kappa) \in \Theta$ as a bi-dimensional hyperparameter to be searched, and formalize the search as a discrete multi-objective optimization problem over complete re-solutions of Algorithm~\ref{alg:repair-lot-skyline}.

\textbf{Definition 11 (Configuration objective vector).} For a finite grid $\Theta = \Theta_{L} \times \Theta_{\kappa}$ of candidate values and a fixed dataset, every $\theta \in \Theta$ is evaluated by re-running Algorithm~\ref{alg:repair-lot-skyline} in full (all other parameters held at the values of Table~\ref{tab:parameters}) to obtain a repair plan $\{L_1, \ldots, L_{M(\theta)}\}$, from which we extract the objective vector
\begin{equation}
\begin{aligned}
o(\theta) = {} & \Big(\, \underbrace{\textstyle\sum_{m} \ell(L_m)}_{\Lambda(\theta)\ \text{(budget, minimize)}},\ \ \underbrace{M(\theta)}_{\text{mobilization count, minimize}},\\
& \underbrace{\tfrac{\text{crack units covered}}{\text{crack units total}}}_{\mathrm{Cov}(\theta)\ \text{(benefit, maximize)}},\ \ \underbrace{\overline{\mathrm{Specificity}}(\theta)}_{\text{mean over final lots, maximize}} \,\Big),
\end{aligned}
\label{eq:config-objective}
\end{equation}
where $\ell(L_m)$ is the real chainage length of final lot $L_m$, $M(\theta) = |\{L_1, \ldots, L_{M(\theta)}\}|$, and $\overline{\mathrm{Specificity}}(\theta)$ is $\mathrm{Specificity}_j(\kappa)$ (Eq.~\ref{eq:patch-specificity}) averaged over the final merged lots of $\theta$. Pothole coverage is not included as an objective: it is identically $1$ for every $\theta$, since pothole is a hard constraint (Eq.~\ref{eq:decision}) independent of $L_{\min}$ and $\kappa$.

\textbf{The mobilization-count guardrail.} An earlier formulation of $o(\theta)$ omitted $M(\theta)$, retaining only budget, crack coverage, and specificity. Because the cost model (Eq.~\ref{eq:benefit-cost-generalized}) is linear in lot length with no fixed per-mobilization term, this reduced objective vector is degenerate: shrinking $L_{\min}$ simultaneously lowers $\Lambda(\theta)$ \emph{and} raises $\mathrm{Cov}(\theta)$ without bound, collapsing every hazard cluster into its own minimal, separately mobilized lot and silently defeating $L_{\min}$'s physical purpose (Section~\ref{sec:minlot}): avoiding a crew mobilization for a single, isolated defect. Including $M(\theta)$ as a fourth, minimized objective restores a genuine trade-off, since consolidating nearby clusters into fewer, longer lots (large $L_{\min}$) and splitting them into many short, separately mobilized lots (small $L_{\min}$) now compete directly.

\textbf{Definition 12 (Pareto dominance and recommendation).} Configuration $\theta$ \emph{Pareto-dominates} $\theta'$ iff $\theta$ is at least as good as $\theta'$ on every coordinate of Eq.~\ref{eq:config-objective} (after orienting all four to a common ``smaller is better'' scale) and strictly better on at least one. The search reports the Pareto-nondominated subset $\Theta^\star \subseteq \Theta$. To select one recommended configuration from $\Theta^\star$ without introducing additional scalar weights, each objective is min--max normalized over the \emph{entire} evaluated grid $\Theta$ (not only $\Theta^\star$) and oriented so that $1$ denotes the best observed value; the recommended configuration is the element of $\Theta^\star$ at minimum Euclidean distance from the all-ones ideal (``utopia'') point in this normalized space---a standard, weight-free knee-point selection on a discrete Pareto frontier.

\subsection{Assumptions and Threats}

The hazard-clustering procedure (Definition 4) assumes that chainage-adjacent crack/pothole observations within $g$ metres represent the same distress area rather than independent defects; $g$ is calibrated from the empirical gap distribution of this specific dataset (Appendix~\ref{sec:appendix-gap}) and would need recalibration for a survey with a different sampling pitch. The soft-constraint threshold (Eq.~\ref{eq:benefit-cost-generalized}--\ref{eq:decision}) encodes a specific, heuristic cost/benefit trade-off rather than a learned or economically calibrated unit repair cost; Section~\ref{sec:skyline-def}'s skyline frontier is precisely the mechanism by which this heuristic threshold can be audited and, if needed, replaced with a different operating point without re-running the clustering and merging steps. The minimum-lot-length expansion (Definition 5) assumes the surveyed route segment is at least $L_{\min}$ long; for a route shorter than $L_{\min}$, the returned lot spans the entire available route and only approximately satisfies the minimum-length constraint.

\section{Experimental Setup}
\label{sec:experiments}

\subsection{Dataset Description}

We apply the Repair Lot Skyline WCSP to the PaveDistress image collection~\citep{PaveDistress2024}, a publicly released (CC BY 4.0) pavement distress survey of the S315 highway, China, collected by a pavement inspection vehicle equipped with a Basler raL2048-80km line-scan camera and infrared laser-assisted lighting. Each image has a resolution of $3854 \times 2065$ pixels and covers an actual pavement area of approximately $3.9 \times 2.1$~m, consistent with the unit-image footprint of Section~\ref{sec:notation}. Images are organized into five top-level categories: \emph{Background} (no visible distress), \emph{Crack} (three subfolders: longitudinal, map, and transverse cracking), \emph{Patch} (completed repair), \emph{Pothole}, and \emph{Other} (ambiguous images, neither confirmed distress nor confirmed sound pavement). Each image's filename encodes its estimated chainage position (in kilometres) along the surveyed highway, which we use directly as the spatial coordinate $c^{(i)}$.

The raw collection contains 6{,}032 files (2{,}784 Background; 824 Crack, split 539/184/101 across longitudinal/map/transverse; 956 Other; 1{,}205 Patch; 263 Pothole), distributed across \texttt{.jpg} and \texttt{.png} extensions. Within Crack, Patch, and Pothole, a base filename appearing with both extensions is a byte-identical duplicate (verified via MD5 hashing), so only the \texttt{.jpg} copy is retained; Background has no overlapping base names between its \texttt{.jpg} and \texttt{.png} files, so both are retained as distinct images. After this deduplication, $n = 4{,}922$ unique unit images remain: 2{,}783 Background, 956 Other, 604 Patch, 441 Crack, and 138 Pothole. Unlike a related, out-of-scope state-transition data pipeline for Markov deterioration hazard modeling that discards the Other category because it has no well-defined ordinal severity mapping~\citep{Yasuno2026Scaling}, this paper's method includes Other directly, mapping it to hazard class \texttt{none} alongside Background (Definition 1), since both represent ``no actionable distress at this location'' for repair-planning purposes and this maximizes the spatial coverage of the hazard map.

\subsection{A Single-Epoch, Spatially Sparse Survey}
\label{sec:sparsity}

Unlike a longitudinal inspection record with repeated visits to the same location, PaveDistress provides exactly one observation per surveyed chainage position: it is a single-epoch spatial snapshot, not a time series, which is precisely why we formulate the planning problem over chainage (Section~\ref{sec:methodology}) rather than attempting any deterioration forecast (Section~\ref{sec:related}).

The nominal chainage range spanned by filenames is $[0.002, 1540.725]$~km. However, applying the data-run detection procedure of Section~\ref{sec:methodology} (grouping non-empty 1~km hazard bins within a 5~km gap tolerance, padded by 2~km) reveals that only four disjoint chainage intervals actually contain surveyed images: $[0, 6]$~km, $[11, 19]$~km, $[466, 490]$~km, and $[1531, 1541]$~km, totaling 48~km---only 3.1\% of the nominal 1,541~km range. This sparsity has a direct, safety-relevant consequence: of the 263 raw Pothole files, 194 (74\%) fall in chainage ranges beyond an initially assumed $\sim$486~km analysis window (corresponding to the third interval above). Because Pothole observations are the hard-constraint trigger of this method (Section~\ref{sec:decision}), truncating the analysis to the initially assumed window would have discarded three-quarters of the available pothole evidence. We therefore retain the full observed chainage range for both the hazard map and the WCSP, a decision confirmed with the project's domain-expert stakeholder before running the experiments reported in Section~\ref{sec:results}.

\subsection{Parameters}
\label{sec:parameters}

Table~\ref{tab:parameters} lists the fixed parameters used for all results in Section~\ref{sec:results}. The gap tolerance $g = 11$~m for hazard clustering (Section~\ref{sec:clustering}) is calibrated from the empirical inter-observation chainage-gap distribution of this dataset: across all categories, the median consecutive-sample gap is 2.07~m and approximately 90\% of gaps are at most 11~m (five unit-image pitches), so $g = 11$~m bridges typical sampling gaps without merging genuinely separate distress areas. The cost weight $w_{\text{cost}} = 0.12$ (Section~\ref{sec:decision}) is calibrated so that a single isolated crack observation, padded to the minimum lot length $L_{\min} = 10.5$~m, is excluded (benefit $1 <$ cost $1.26$), while any cluster with two or more nearby crack observations, or one already spanning more than the minimum length, is included---operationalizing the qualitative instruction that crack repair should be pursued ``as much as cost-justified'' rather than for an isolated, low-confidence single observation. The primary case study of Section~\ref{sec:results} sets $w_{\text{rec}} = 0$ (the baseline model, Eq.~\ref{eq:benefit-cost}); Section~\ref{sec:hyperparam-experiments} instead activates $w_{\text{rec}} = 0.2$ for the hyperparameter grid search, since $w_{\text{rec}} = 0$ would make $\kappa$ decision-irrelevant and render a search over it vacuous (Section~\ref{sec:hyperparam-opt}).

\begin{table}[h]
\centering
\caption{Fixed parameters used in all reported experiments (Section~\ref{sec:results}); the hyperparameter grid search of Section~\ref{sec:hyperparam-experiments} instead sweeps $L_{\min}, \kappa$ and additionally sets $w_{\text{rec}} = 0.2$, as noted above.}
\label{tab:parameters}
\begin{tabular}{llr}
\toprule
\textbf{Symbol} & \textbf{Description} & \textbf{Value} \\
\midrule
$w$ & Unit-image lane width & 3.9~m \\
$p$ & Unit-image chainage pitch & 2.1~m \\
$b$ & Kilopost hazard-bin width & 1~km \\
$g$ & Hazard-cluster chainage gap tolerance & 11~m \\
$u_{\min}$ & Minimum lot length, in pitch units & 5 \\
$L_{\min}$ & Minimum serviceable lot length ($p \cdot u_{\min}$) & 10.5~m \\
$\kappa$ & Patch historical-context radius & 50~m \\
$\kappa_0$ & Specificity reference radius (Eq.~\ref{eq:patch-specificity}) & 15~m \\
$w_{\text{soft}}$ & Soft-constraint (crack) benefit weight & 1.0 \\
$w_{\text{rec}}$ & Soft-constraint recurrence-risk benefit weight & 0 (primary study); 0.2 (grid search) \\
$w_{\text{cost}}$ & Soft-constraint cost weight per metre & 0.12 \\
\bottomrule
\end{tabular}
\end{table}

\subsection{Hyperparameter Grid-Search Setup}
\label{sec:hyperparam-experiments}

To evaluate the hyperparameter optimization of Section~\ref{sec:hyperparam-opt} on real data, we define two grids. The minimum-lot-length grid is
\begin{equation}
\Theta_L = \{6.3,\ 10.5,\ 14.7,\ 21.0,\ 31.5,\ 42.0\}~\text{m},
\label{eq:theta-l-grid}
\end{equation}
the integer multiples $u_{\min} \in \{3, 5, 7, 10, 15, 20\}$ of the unit-image pitch $p = 2.1$~m, so that every candidate $L_{\min}$ remains a physically meaningful ``$N$ unit images'' length ($10.5$~m is the Table~\ref{tab:parameters} default). The historical-context-radius grid is
\begin{equation}
\Theta_\kappa = \{10,\ 25,\ 50,\ 75,\ 100,\ 150\}~\text{m},
\label{eq:theta-kappa-grid}
\end{equation}
bracketing the Table~\ref{tab:parameters} default of $50$~m. Together these give $|\Theta_L \times \Theta_\kappa| = 36$ configurations. Every other parameter is held at its Table~\ref{tab:parameters} value, including $w_{\text{rec}} = 0.2$. Algorithm~\ref{alg:repair-lot-skyline} is re-run once per configuration (no warm-starting or memoization between configurations); on the case-study dataset of Section~\ref{sec:experiments}, all 36 configurations solve in approximately 3 seconds of wall-clock time on a single consumer workstation, consistent with the $O(n \log n)$ per-configuration complexity of Appendix~\ref{sec:appendix-methodology}.

\subsection{Implementation and Software}

\begin{itemize}
    \item \textbf{Programming language}: Python 3.13
    \item \textbf{Data handling}: pandas 2.2.3~\citep{McKinney2010}, NumPy 2.3.5~\citep{Harris2020}
    \item \textbf{Visualization}: Matplotlib 3.10.8~\citep{Hunter2007} with the \texttt{matplotlib-fontja} extension for mixed Japanese/English figure text
    \item \textbf{Z-score hazard scaling} (Eq.~\ref{eq:zscore-hazard}): a shared internal scaling utility also used, independently, by the out-of-scope deterioration-modeling study~\citep{Yasuno2026Scaling}; only z-score scaling is used here, with no alternative-method comparison
    \item \textbf{WCSP implementation}: a dedicated internal Python module implementing hazard clustering, minimum-lot-length expansion, patch-history attachment, the hard/soft decision rule, lot merging, and skyline frontier construction (Section~\ref{sec:methodology}); retained internally and not released as part of this paper
    \item \textbf{Testing}: an automated unit-test suite covering cluster gap-splitting, minimum-lot-length padding (including route-boundary clipping), patch-recurrence counting, hard/soft decision correctness, lot merging, and skyline frontier monotonicity
\end{itemize}

\subsection{Reproducibility}

Unlike the random-seed-dependent state-transition pipeline used for the related deterioration-modeling study, the Repair Lot Skyline WCSP of this paper is entirely deterministic given the fixed parameters of Table~\ref{tab:parameters}: hazard clustering, minimum-lot-length expansion, the hard/soft decision rule, lot merging, and skyline frontier construction involve no random sampling. The same procedure applied to the same extracted PaveDistress folder therefore reproduces byte-identical candidate-lot, repair-plan, and skyline-frontier tables. Code and the case-study dataset are retained internally for confidentiality and are not distributed externally as part of this paper; the dataset itself is separately available under CC BY 4.0 from its original publisher~\citep{PaveDistress2024}.

\section{Results}
\label{sec:results}

\subsection{Distress Composition by Survey Segment}

Table~\ref{tab:segments} reports the hazard-class composition of each of the four disjoint surveyed chainage segments identified in Section~\ref{sec:sparsity}. Distress composition varies markedly by segment: the $[466, 490]$~km segment is dominated by \texttt{none} (3{,}137 of 3{,}693 unit images, 84.9\%) and \texttt{patch} (485, 13.1\%) observations, with comparatively few crack (56) and pothole (15) observations, whereas the $[1531, 1541]$~km segment---only 186 unit images, the smallest segment---contains the highest pothole density of any segment (91 observations, 48.9\% of that segment's images). This segment-level heterogeneity is precisely why a uniform, dataset-wide severity threshold would be a poor substitute for the spatial clustering and per-segment hard/soft decision applied in this paper (Section~\ref{sec:methodology}).

\begin{table}[h]
\centering
\caption{Unit-image hazard-class composition by surveyed chainage segment ($n = 4{,}922$ total).}
\label{tab:segments}
\begin{tabular}{lrrrrr}
\toprule
\textbf{Segment (km)} & \textbf{Total} & \textbf{None} & \textbf{Crack} & \textbf{Patch} & \textbf{Pothole} \\
\midrule
$[0, 6]$     & 729   & 602   & 89  & 8   & 30 \\
$[11, 19]$   & 314   & 0     & 273 & 39  & 2  \\
$[466, 490]$ & 3{,}693 & 3{,}137 & 56  & 485 & 15 \\
$[1531, 1541]$ & 186 & 0     & 23  & 72  & 91 \\
\midrule
\textbf{Total} & \textbf{4{,}922} & \textbf{3{,}739} & \textbf{441} & \textbf{604} & \textbf{138} \\
\bottomrule
\end{tabular}
\end{table}

\begin{figure}[h]
\centering
\includegraphics[width=\textwidth]{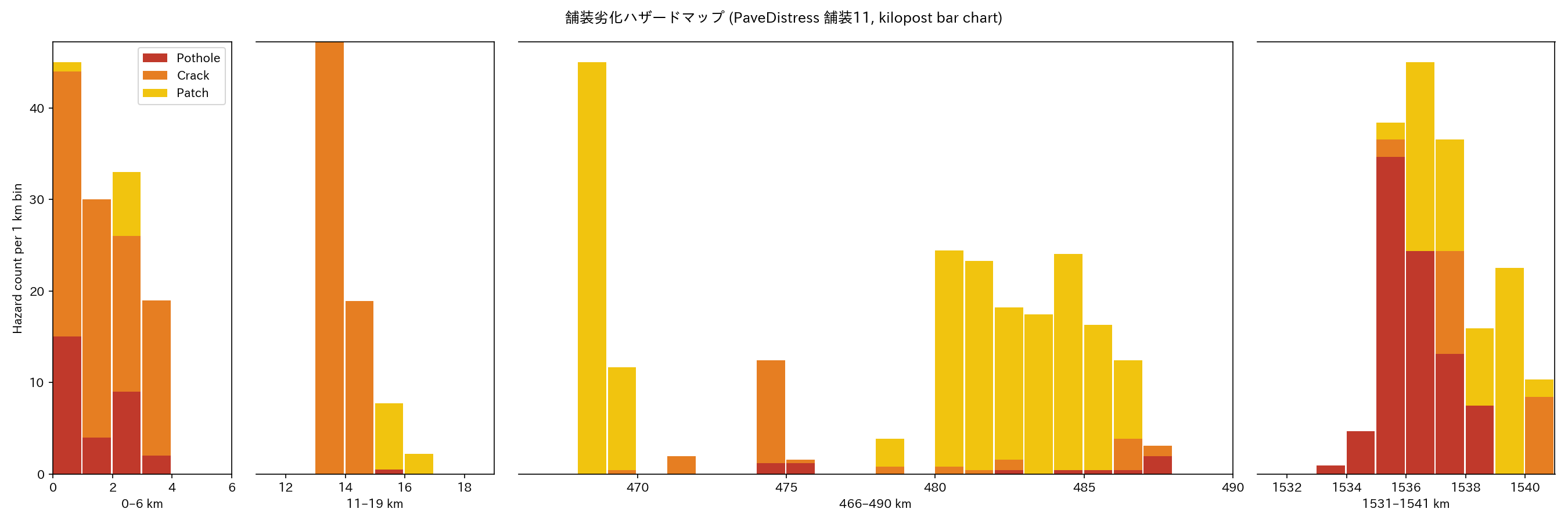}
\caption{Pavement degradation hazard density (pothole, crack, patch counts per 1~km kilopost bin), shown only for the four chainage segments actually covered by the survey. Panel widths are proportional to each segment's chainage span (best viewed in color).}
\label{fig:hazard-barchart}
\end{figure}

Figure~\ref{fig:hazard-barchart} visualizes this same composition as a kilopost-level stacked hazard-density bar chart (pothole, crack, patch counts per 1~km bin), restricted to the four surveyed segments via the data-run detection procedure of Section~\ref{sec:methodology}; the intervening $\sim$1,493~km of nominal chainage with no survey coverage is omitted from the chart rather than rendered as a long run of empty bins.

\subsection{Hazard Clusters and WCSP Decisions}

Applying hazard clustering (Section~\ref{sec:clustering}) to the 579 crack/pothole trigger observations ($441$ crack $+$ $138$ pothole) yields $J = 143$ candidate clusters: 61 hard (pothole-containing) and 82 soft (crack-only). Table~\ref{tab:wcsp-outcome} summarizes the resulting WCSP decisions. All 61 hard clusters are included unconditionally (Eq.~\ref{eq:decision}); of the 82 soft clusters, 45 (54.9\%) satisfy the benefit-to-cost threshold and are included, while 37 (45.1\%) are excluded. Merging overlapping or touching included lots (Section~\ref{sec:methodology}) produces $M = 106$ final repair lots---identical in count to the 106 included candidate clusters, indicating that for this dataset and parameter configuration, no two included clusters were close enough after minimum-length padding to require merging into a single lot. The final repair plan totals 1{,}997.4~m, with every lot satisfying the 10.5~m minimum-length constraint by construction (minimum observed lot length: 10.5~m; maximum: 187.9~m; mean: 18.8~m).

\begin{table}[h]
\centering
\caption{Repair Lot Skyline WCSP outcome summary.}
\label{tab:wcsp-outcome}
\begin{tabular}{lr}
\toprule
\textbf{Quantity} & \textbf{Value} \\
\midrule
Candidate hazard clusters (total) & 143 \\
\quad Hard (pothole-containing) & 61 \\
\quad Soft (crack-only) & 82 \\
Soft clusters included (benefit $\geq$ cost) & 45 \\
Soft clusters excluded (benefit $<$ cost) & 37 \\
Final merged repair lots & 106 \\
Total repair-plan length & 1{,}997.4~m \\
Pothole coverage & 138 / 138 (100.0\%) \\
Crack coverage & 404 / 441 (91.6\%) \\
Lots with nonzero patch-history annotation & 14 / 106 \\
\bottomrule
\end{tabular}
\end{table}

\begin{figure}[h]
\centering
\includegraphics[width=\textwidth]{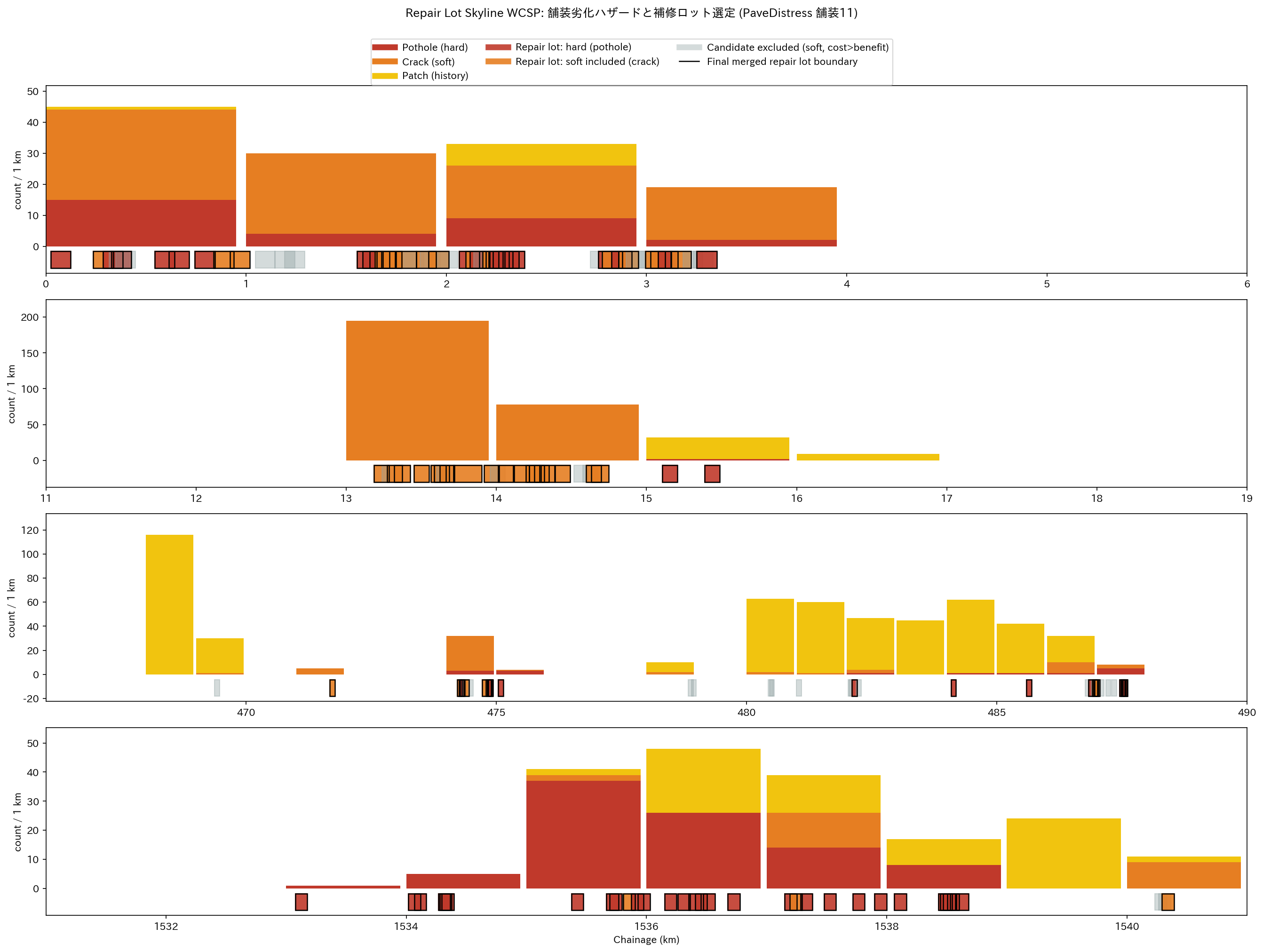}
\caption{Repair Lot Skyline: kilopost hazard density (top, per segment row) with the WCSP repair-lot bands (hard/soft-included/soft-excluded) and final merged repair-lot boundaries (black outline) overlaid beneath (best viewed in color).}
\label{fig:repair-lot-skyline}
\end{figure}

Figure~\ref{fig:repair-lot-skyline} overlays the WCSP outcome on the kilopost hazard-density chart: hard lots (red), included soft lots (orange), and excluded soft candidates (grey) are shown as bands beneath the bars, with the final merged repair-lot boundaries outlined in black. Because the pothole hard constraint forces inclusion regardless of local crack density, hard lots are visible even in otherwise low-hazard bins (e.g., within the $[1531,1541]$~km segment, where pothole density is highest), while excluded soft candidates cluster in regions of low, isolated crack density (e.g., within the sparser portions of $[466,490]$~km).

\subsection{Skyline Frontier: Budget versus Coverage}

\begin{figure}[h]
\centering
\includegraphics[width=0.765\textwidth]{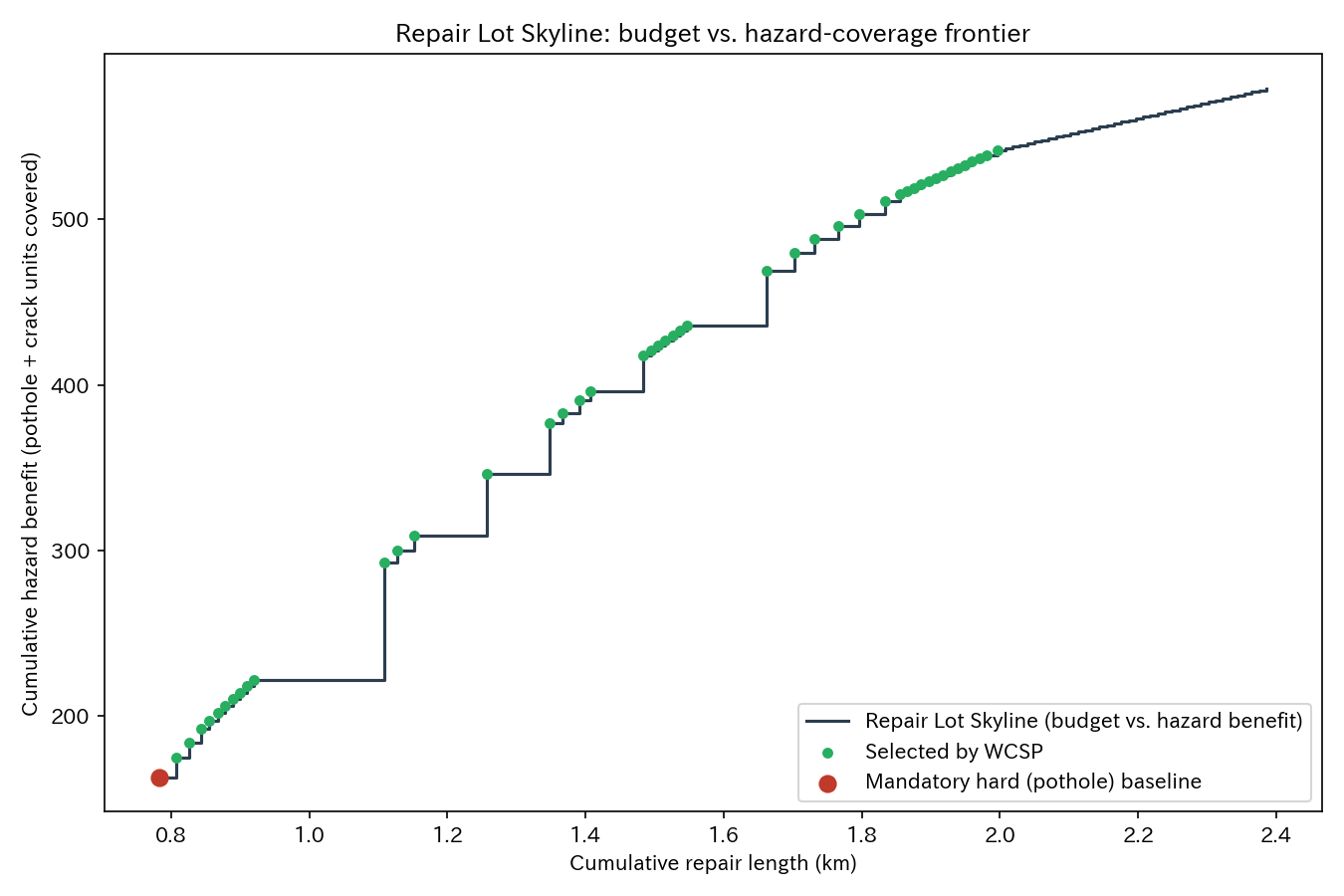}
\caption{Skyline frontier: cumulative repair length (budget) versus cumulative hazard benefit (coverage). The red point marks the mandatory hard (pothole) baseline; green points mark soft candidates selected by the WCSP's default cost/benefit threshold; the frontier continues (unselected) toward the full-budget point where every candidate is included regardless of cost (best viewed in color).}
\label{fig:skyline-frontier}
\end{figure}

Figure~\ref{fig:skyline-frontier} plots the cumulative skyline frontier of Definition 10: cumulative repair length (budget) on the horizontal axis against cumulative hazard benefit (pothole and crack observations covered) on the vertical axis. The mandatory hard baseline alone requires 783.2~m (39.2\% of the final selected budget) and covers a benefit of 163 (30.1\% of the final selected benefit)---that is, the pothole hard constraint by itself accounts for less than 40\% of the repair-plan length but only about 30\% of total covered benefit, because pothole clusters are comparatively short and localized relative to the cumulative length contributed by included crack clusters.

The WCSP's default cost/benefit threshold selects an operating point at 1{,}997.4~m, capturing 542 of the 579 total addressable benefit units in the full candidate set (93.6\%) using only 83.7\% of the 2{,}385.9~m that would be required to include every soft candidate regardless of cost. Beyond this point, the frontier flattens sharply: the remaining 37 excluded soft clusters, if all included, would add only 37 further benefit units (6.8\% more than the selected benefit) at a cost of 388.5~m of additional repair length (19.4\% more than the selected budget)---a benefit-per-metre efficiency roughly one-third of the already-included soft clusters. This diminishing-returns pattern is consistent with the efficiency-descending construction of the frontier (Definition 9--10) and provides the quantitative justification for treating the default threshold as a reasonable, though not unique, operating point: a planner with a smaller budget could move left along the frontier (accepting less crack coverage while still covering 100\% of potholes, since the hard baseline is always the leftmost feasible point), while a planner with additional budget could move right, with precisely quantified diminishing returns.

\subsection{Hyperparameter Optimization Results}
\label{sec:hyperparam-results}

\begin{figure}[h]
\centering
\includegraphics[width=0.765\textwidth]{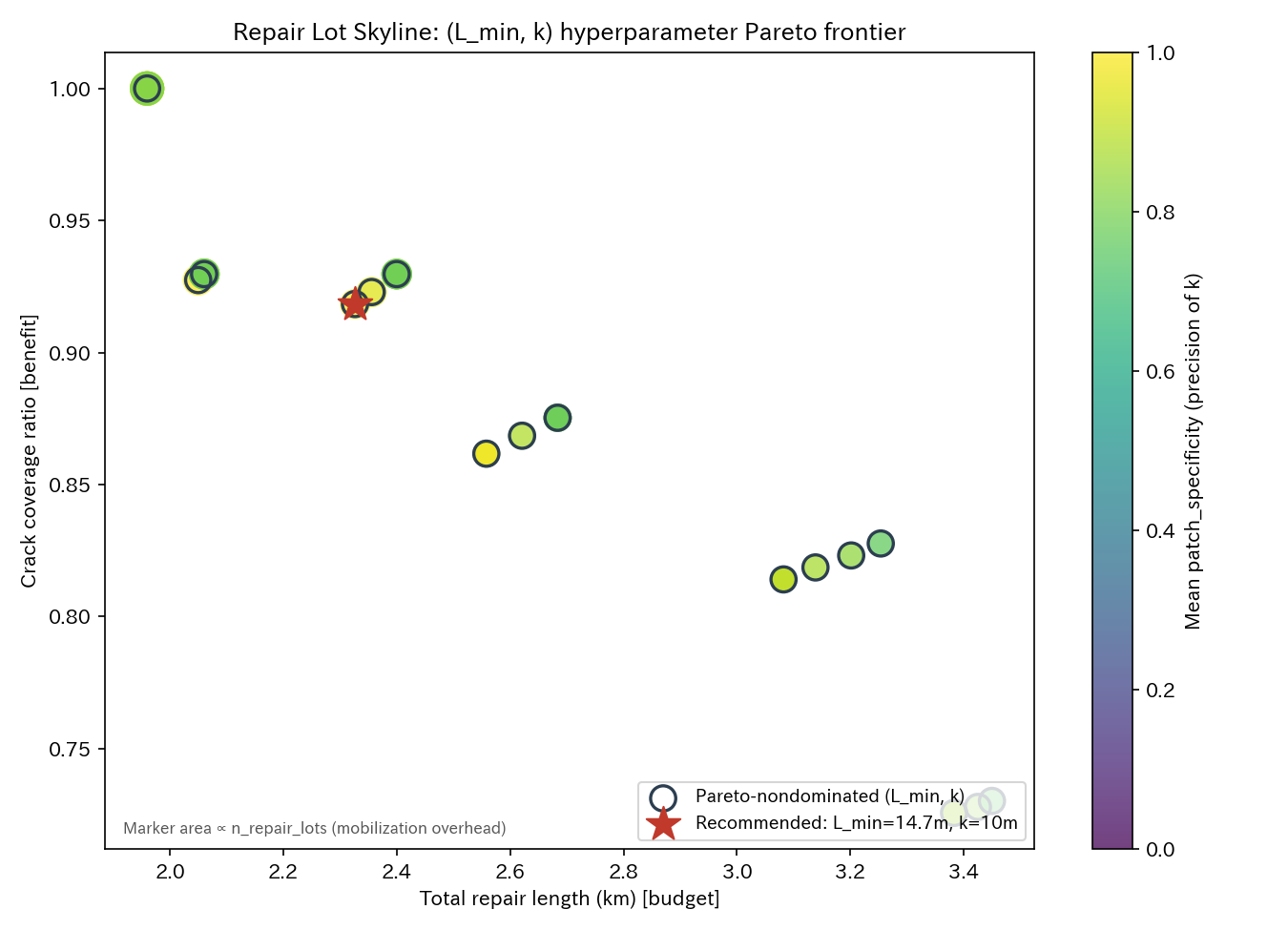}
\caption{Hyperparameter Pareto frontier over $(L_{\min}, \kappa)$: every evaluated configuration in the cumulative-budget-versus-crack-coverage plane. Color encodes mean patch specificity $\overline{\mathrm{Specificity}}(\theta)$; marker area is proportional to mobilization count $M(\theta)$; hollow rings mark Pareto-nondominated configurations $\Theta^\star$; the star marks the recommended configuration (best viewed in color).}
\label{fig:hyperparam-pareto}
\end{figure}

Applying the $(L_{\min}, \kappa)$ grid search of Section~\ref{sec:hyperparam-experiments} to the same dataset yields, of the $|\Theta| = 36$ evaluated configurations, $|\Theta^\star| = 16$ Pareto-nondominated configurations under the four-objective vector of Eq.~\ref{eq:config-objective}. Figure~\ref{fig:hyperparam-pareto} plots every evaluated configuration in the (budget, crack-coverage) plane, with point color encoding mean patch specificity and marker area encoding mobilization count $M(\theta)$; hollow rings mark $\Theta^\star$ and the star marks the recommended configuration selected by the ideal-point rule of Definition 12: $L_{\min} = 14.7$~m ($u_{\min} = 7$ unit images), $\kappa = 10$~m.

Table~\ref{tab:hyperparam-comparison} compares the recommended configuration against the Table~\ref{tab:parameters} default ($L_{\min} = 10.5$~m, $\kappa = 50$~m), both evaluated with $w_{\text{rec}} = 0.2$ active so the comparison isolates the effect of $(L_{\min}, \kappa)$ alone. Relative to the default, the recommended configuration reduces mobilization count by 7.1\% (8 fewer repair lots) at the cost of a 12.9\% larger repair-length budget and a 1.1-percentage-point reduction in crack coverage, while increasing mean patch specificity from 0.874 to a perfect 1.0.

\begin{table}[h]
\centering
\caption{Default versus recommended $(L_{\min}, \kappa)$ configuration, both with $w_{\text{rec}} = 0.2$.}
\label{tab:hyperparam-comparison}
\begin{tabular}{lrrr}
\toprule
\textbf{Quantity} & \textbf{Default} & \textbf{Recommended} & \textbf{Change} \\
 & $(L_{\min}{=}10.5\text{m}, \kappa{=}50\text{m})$ & $(L_{\min}{=}14.7\text{m}, \kappa{=}10\text{m})$ & \\
\midrule
Final merged repair lots $M(\theta)$ & 112 & 104 & $-7.1\%$ \\
Total repair-plan length $\Lambda(\theta)$ & 2{,}060.4~m & 2{,}326.4~m & $+12.9\%$ \\
Pothole coverage & 138/138 (100.0\%) & 138/138 (100.0\%) & $0.0$ pt \\
Crack coverage $\mathrm{Cov}(\theta)$ & 410/441 (93.0\%) & 405/441 (91.8\%) & $-1.1$ pt \\
Mean patch specificity $\overline{\mathrm{Specificity}}(\theta)$ & 0.874 & 1.000 & $+0.126$ \\
\bottomrule
\end{tabular}
\end{table}

Two trends are visible across the full grid (Figure~\ref{fig:hyperparam-pareto} and the underlying 36-row table). First, holding $\kappa = 50$~m fixed and sweeping $L_{\min}$ across $\Theta_L$ traces a monotone mobilization-versus-budget-versus-coverage trade-off: $M(\theta)$ falls from 143 to 61 lots (a 57\% reduction) as $L_{\min}$ increases from 6.3~m to 42.0~m, while $\Lambda(\theta)$ rises from 1{,}959.8~m to 3{,}382.9~m ($+73\%$) and $\mathrm{Cov}(\theta)$ falls from 100.0\% to 72.6\% ($-27.4$ pt)---precisely the trade-off the mobilization-count guardrail of Section~\ref{sec:hyperparam-opt} was designed to expose. Second, holding $L_{\min} = 10.5$~m fixed and sweeping $\kappa$ across $\Theta_\kappa$ shows pronounced diminishing returns: $\mathrm{Cov}(\theta)$ moves only from 409 to 410 covered crack units (a single additional lot) across the entire range $\kappa \in [10, 150]$~m, with essentially all of that gain already realized by $\kappa = 25$~m, while $\overline{\mathrm{Specificity}}(\theta)$ falls monotonically from 1.0 ($\kappa = 10$~m) to 0.773 ($\kappa = 150$~m)---a 22.7-percentage-point loss in recurrence-signal precision for a coverage gain of at most one lot. This asymmetry is why the recommended configuration selects a small $\kappa$ (10~m) rather than the Table~\ref{tab:parameters} default (50~m): once $L_{\min}$ is also being tuned, a narrow $\kappa$ captures essentially all of the achievable crack-coverage benefit from the recurrence-risk bonus while keeping the recurrence signal maximally specific.

\section{Discussion}
\label{sec:discussion}

\subsection{Why Heterogeneous Constraint Semantics Matter}

A single severity-ranking approach to repair-lot selection would implicitly treat pothole, crack, and patch evidence as differing only in degree, to be thresholded by one scalar cutoff. Our results show why this would be a poor model of the actual decision problem. The hard pothole constraint accounts for only 39.2\% of the final repair budget but a disproportionately lower 30.1\% of covered benefit (Section~\ref{sec:results}), precisely because pothole clusters are short and localized: a severity-threshold model calibrated to maximize aggregate benefit-per-cost would, if given no safety override, systematically under-prioritize these short, low-aggregate-benefit but safety-critical locations relative to longer, higher-aggregate-benefit crack regions. The explicit hard constraint of Eq.~\ref{eq:decision} removes this trade-off entirely for potholes, by design, rather than relying on a severity score large enough to dominate the ranking in practice.

Similarly, the historical patch-recurrence annotation (Section~\ref{sec:patch-history}) is deliberately non-binding: only 14 of 106 final lots (13.2\%) carry a nonzero patch-history count (Table~\ref{tab:wcsp-outcome}), meaning most repair decisions in this case study are made without prior-repair context. Had patch history instead been folded into the benefit score as an additional severity signal, it would have altered soft-constraint inclusion decisions at locations where no current crack or pothole evidence exists---effectively re-triggering a repair based on history alone, contrary to the problem statement that re-degradation risk is contextual rather than actionable on its own (Section~\ref{sec:introduction}).

\subsection{Hazard Condition Reality: Environmental and Logistics Context for the Hard/Soft Rule}
\label{sec:hazard-reality}

The hard/soft decision rule of Section~\ref{sec:decision} is not merely a modeling convenience; it is motivated by operating conditions characteristic of this class of logistics corridor, even though we do not have access to site-specific climate or traffic-volume instrumentation for the S315 highway itself. Four such considerations are worth making explicit, and they argue for, rather than against, treating pothole and crack asymmetrically.

First, highways in cold, high-rainfall regions are subject to severe climatic loading: substantial seasonal rainfall together with winter conditions in which standing surface moisture repeatedly freezes is a combination known in pavement engineering to accelerate the transition from a surface crack to a structural pothole far faster than dry or temperate conditions would. Second, winter maintenance in snow-affected regions commonly applies deicing salt, whose dissolved chloride ions lower the local freezing point of residual moisture within surface cracks. The resulting repeated freeze--thaw cycling of water trapped inside the pavement structure is a well-documented physical mechanism for pothole initiation and growth, distinct from the slower, load-driven propagation that dominates ordinary crack widening. Third, a highway of this class typically functions as a logistics lifeline, carrying a high proportion of heavy freight traffic. Freight-vehicle axle loads convert a freeze--thaw-weakened crack into a safety-critical pothole far more readily than passenger-car traffic would, and drivers of such vehicles have comparatively little ability to swerve around a sudden pavement failure at highway speed. Fourth, and specific to this corridor's economic role, the chainage span between one interchange's entrance and the next interchange's exit functions as a logistics \emph{meta-path}. It is a single, named segment of highway that stands in for the physical linkage between distribution warehouses and manufacturing sites that are not themselves located directly on the surveyed road but are accessed through it. A pothole anywhere along such a meta-path segment is therefore a hazard to every shipment traversing that link, not only to traffic destined for the segment itself. This is precisely the kind of disproportionate, safety- and continuity-critical consequence that justifies a hard, unconditional inclusion rule (Eq.~\ref{eq:decision}), rather than a cost/benefit threshold calibrated only against the repair budget. Cracks, by contrast, degrade ride quality and accelerate future deterioration but do not typically produce the same abrupt, vehicle-damaging failure mode under heavy axle loads, which is consistent with treating them as a soft, discretionary constraint whose inclusion can reasonably depend on available budget (Section~\ref{sec:decision}).

\subsection{The Skyline Frontier Reveals Diminishing Returns, Not Just a Threshold}

The cumulative skyline frontier (Figure~\ref{fig:skyline-frontier}) shows that the WCSP's default threshold already captures 93.6\% of total addressable benefit using 83.7\% of the maximum conceivable budget, and that the remaining soft candidates have markedly lower benefit-per-metre efficiency than those already included (Section~\ref{sec:results}). This is a direct consequence of sorting candidates by efficiency before accumulation (Definition 9--10): the frontier is, by construction, concave in the sense that each additional increment of budget buys no more benefit than the previous increment. The practical value of exposing this curve, rather than only the single selected plan, is that a planner facing a tighter or looser budget than assumed by the default cost weight $w_{\text{cost}}$ can read off the achievable trade-off directly, without re-running the clustering and decision procedure with a different parameter value for every candidate budget of interest.

\subsection{Hyperparameter Sensitivity: Mobilization Count as a Guardrail Objective}
\label{sec:discussion-hyperparam}

The hyperparameter search of Section~\ref{sec:hyperparam-results} surfaces a methodological pitfall that is easy to miss when a cost model is linear in length alone (Eq.~\ref{eq:benefit-cost-generalized}): without a mobilization-count objective, the search degenerates to a single trivial optimum at the smallest evaluated $L_{\min}$, which simultaneously minimizes budget, maximizes crack coverage, and maximizes patch specificity by fragmenting every hazard cluster into its own separately mobilized lot (Section~\ref{sec:hyperparam-opt}). This is not a sign that small $L_{\min}$ is actually preferable; it is an artifact of an incomplete objective vector that omits the real-world cost of dispatching a repair crew to many scattered, minimal-length locations rather than fewer, consolidated ones. Once $M(\theta)$ is reinstated as a fourth, minimized objective, the Pareto frontier becomes non-trivial (16 of 36 configurations nondominated; Figure~\ref{fig:hyperparam-pareto}), and $L_{\min}$ resumes its intended role as a genuine mobilization-versus-coverage dial rather than a parameter with a spuriously dominant setting. We flag this as a general lesson for any WCSP- or skyline-style repair-planning formulation with a purely length-proportional cost: any hyperparameter that can shrink the \emph{unit of consolidation} (here, the lot) must be paired with an explicit penalty on the \emph{number} of consolidated units, or a search over it will silently reward fragmentation.

A second, independent finding concerns the historical-context radius $\kappa$: the Table~\ref{tab:hyperparam-comparison} recommended configuration selects $\kappa = 10$~m rather than the Table~\ref{tab:parameters} default of $50$~m, because crack coverage gains from widening $\kappa$ saturate almost immediately (by $\kappa \approx 25$~m in this dataset) while mean patch specificity continues to fall across the entire swept range (Section~\ref{sec:hyperparam-results}). This is a direct empirical confirmation of the monotonicity argued in Definition 6: $\kappa$ trades recall (more patches counted, marginally more soft clusters tipped into inclusion) against precision (what fraction of counted patches are genuinely local), and the trade-off is asymmetric in this dataset---recall saturates quickly, precision keeps eroding. A practitioner tuning $\kappa$ in isolation, without also tracking $\overline{\mathrm{Specificity}}(\theta)$, would see no apparent downside to widening it indefinitely; the specificity objective is what makes the cost of doing so visible.

\subsection{No Merging Was Needed at This Gap Tolerance}

An incidental finding is that the final repair plan (106 lots) exactly matches the number of included candidate clusters (106), meaning the lot-merging step (Section~\ref{sec:methodology}) performed no actual merges for this dataset and parameter configuration: no two included, minimum-length-padded clusters were close enough to overlap or touch. This is consistent with the chosen cluster gap tolerance $g = 11$~m (Section~\ref{sec:parameters}) already separating distinct distress clusters by construction, combined with the 10.5~m minimum-lot padding (Eq.~\ref{eq:minlot}) not being large enough, at the observed inter-cluster spacing in this dataset, to cause adjacent clusters' padded spans to collide. We retain the merging step in the method specification because it is required for correctness whenever $g$ or $L_{\min}$ are increased, or when applied to a different, more densely clustered survey; its correctness on overlapping synthetic inputs is confirmed by the unit-test suite (Section~\ref{sec:experiments}) even though it was not exercised by this particular case study's real-data run.

\subsection{Practical Implications}

The method is designed to be directly usable by a maintenance planner without requiring a fitted deterioration model, a multi-year planning horizon, or repeated inspections of the same location: it requires only a single distress survey with per-image category labels and chainage coordinates, both of which are commonly produced by existing image-based pavement distress detection pipelines~\citep{Cao2020}. The skyline frontier (Figure~\ref{fig:skyline-frontier}) in particular is intended as a budget-negotiation artifact: it can be presented alongside a specific proposed repair plan to justify why the plan stops where it does, and to quantify the marginal cost of extending it.

\subsection{Limitations}
\label{sec:limitations}

\textbf{Heuristic, uncalibrated cost weight.} The soft-constraint cost weight $w_{\text{cost}} = 0.12$ (Section~\ref{sec:parameters}) is calibrated qualitatively, to produce an intuitively reasonable include/exclude boundary at the scale of one versus several crack observations, rather than estimated from actual unit repair costs or agency budget data. Agencies adopting this method should recalibrate $w_{\text{cost}}$ (or, more generally, replace Eq.~\ref{eq:benefit-cost} with an economically grounded cost model) against their own cost data; the skyline frontier construction itself does not depend on this specific calibration and remains valid under any monotone cost model.

\textbf{No claim about future deterioration.} This method is explicitly restricted to decision support from an already-observed, single-epoch distress inventory (Section~\ref{sec:introduction}); it makes no forecast of which locations will newly develop cracks or potholes, nor any estimate of how quickly an untreated crack might progress to a pothole. A related, separately published line of work addresses Markov deterioration hazard modeling from paired (pseudo-)temporal inspection transitions~\citep{Yasuno2026Scaling}; combining a deterioration forecast with the present repair-lot selection method, so that candidate lots could be prioritized by both current hazard and projected near-term risk, is a natural but unexplored extension. Appendix~\ref{sec:appendix-spatial-hazard} elaborates on how that companion study instead reinterprets its hazard model over chainage, rather than time, on this same dataset, and on a covariate-leakage pitfall encountered and corrected in doing so.

\textbf{Unsurveyed chainage is absent evidence, not absence of hazard.} The data-run restriction of Section~\ref{sec:sparsity} means the method reports on, and only on, the 48~km of chainage actually covered by the survey; the remaining $\sim$1,493~km of nominal highway length is not assessed, and no interpolation or extrapolation of hazard density into unsurveyed gaps is performed or implied. Any deployment of this method should make this scope limitation explicit to decision-makers who might otherwise read the absence of a repair lot as evidence of pavement in good condition, rather than as absent data.

\textbf{Lot-merging behavior untested on this dataset.} As discussed above, the merging step was not exercised on real data in this case study; its correctness rests on unit tests over synthetic overlapping inputs. A denser survey, a larger gap tolerance $g$, or a larger minimum lot length $L_{\min}$ would be expected to produce cases where multiple included clusters are merged into a single final lot, and validating this behavior on a real, densely surveyed network is left to future work.

\textbf{Single-case-study and partial-sensitivity evaluation.} All reported results derive from one dataset (PaveDistress, S315 highway). Section~\ref{sec:hyperparam-results} reports a sensitivity analysis over $L_{\min}$ and $\kappa$ jointly (36 configurations, Figure~\ref{fig:hyperparam-pareto}), but $g$ and $w_{\text{cost}}$ remain fixed at the Table~\ref{tab:parameters} values throughout, and no comparison against an alternative repair-planning heuristic is reported. Extending the multi-objective search of Section~\ref{sec:hyperparam-opt} to include $g$ and $w_{\text{cost}}$ as additional swept dimensions, and evaluating transferability on a second, independently surveyed road network, are natural directions for follow-up work once comparable multi-network pavement distress survey data becomes available.

\section{Conclusion}
\label{sec:conclusion}

We presented the Repair Lot Skyline problem: translating a single-epoch, spatially indexed pavement distress survey into a bounded repair-lot plan by formalizing the heterogeneous roles of pothole (safety-critical, hard constraint), crack (discretionary, soft constraint), and patch (historical, non-binding context) evidence as a Weighted Constraint Satisfaction Problem defined over chainage. The method pads every candidate lot to a minimum serviceable length using real chainage distance, so that it is robust to arbitrarily sparse or irregular spatial sampling, and constructs a cumulative skyline frontier that exposes the full budget-versus-coverage trade-off of discretionary inclusion rather than a single opaque operating point.

Applied to the PaveDistress S315 highway image collection---a public, single-epoch survey whose 4,922 deduplicated unit images cover only 48 of a nominal 1,541~km of chainage across four disjoint segments---the method identifies 143 candidate hazard clusters, selects 106 final repair lots totaling 1,997.4~m, covers 100\% of observed potholes and 91.6\% of observed cracks, and captures 93.6\% of total addressable hazard benefit using 83.7\% of the maximum conceivable repair budget. The skyline frontier shows that the excluded remainder offers markedly diminished returns, providing a quantitative, inspectable basis for budget decisions that a single severity threshold would not surface.

This contribution is deliberately scoped to decision support from already-observed distress, not deterioration forecasting; we view it as complementary to, rather than a replacement for, Markov deterioration hazard modeling approaches that require repeated temporal inspection data. We further formalized the minimum-lot-length $L_{\min}$ and historical-context radius $\kappa$ as a joint, four-objective (budget, mobilization count, crack coverage, patch specificity) hyperparameter search, and found that a mobilization-count guardrail objective is necessary to prevent the search from collapsing to a degenerate, over-fragmented optimum; on the case-study dataset, the recommended configuration ($L_{\min} = 14.7$~m, $\kappa = 10$~m) trades a 12.9\% larger repair budget and a 1.1-percentage-point lower crack coverage for a 7.1\% reduction in repair-crew mobilizations relative to the initial default. Future work includes calibrating the soft-constraint cost weight against real repair-cost data, validating the lot-merging procedure on a more densely surveyed network, combining this spatial decision-support method with a deterioration forecast to incorporate projected near-term risk into lot prioritization, and extending the hyperparameter search to the hazard-cluster gap tolerance $g$ and cost weight $w_{\text{cost}}$ across multiple road networks.

A further, longer-horizon direction concerns the per-image hazard label itself. The present study, like the PaveDistress dataset it uses, takes the per-image damage category ($\texttt{Crack}$, $\texttt{Patch}$, $\texttt{Pothole}$, \emph{etc.}) as already given by prior image classification (Section~\ref{sec:experiments}); it does not address how that label is produced. We see an opportunity to replace this fixed categorical label with a fine-tuned vision--language model (VLM) encoder that takes a damage image as input and generates a free-form textual description of the observed distress, then fine-tune that encoder specifically so that newly captured pavement images are mapped to an accurate damage class alongside the generated text. Unlike a scalar condition index such as the International Roughness Index (IRI) or Maintenance Control Index (MCI), a textual description can express compound, graded, and co-occurring distress characteristics (for example, crack orientation, estimated severity, and surrounding patch history) that a single number necessarily discards. Embedding this text into a high-dimensional (e.g., 1024-dimensional) representation would then let two repair-lot candidates be compared not only by their hazard \emph{counts} (Definition 4) but by the \emph{semantic similarity} of their embedded damage descriptions, opening a path toward \emph{Semantic Repair-Lot Planning}: clustering and prioritizing candidate lots by semantic resemblance to historically severe or rapidly re-degrading locations, rather than by category count alone. We view this as a natural long-term extension of the present WCSP formulation toward a more general \emph{Pavement Repair Intelligence}, in which the spatial decision-support contribution of this paper and a richer, language-model-based perception layer are combined rather than developed in isolation.

\bibliographystyle{unsrtnat}
\bibliography{references}

\appendix
\section{Extended Methodology and Implementation Details}

\subsection{Extended Methodology Notes}
\label{sec:appendix-methodology}

This appendix collects supporting detail for Section~\ref{sec:methodology} and Section~\ref{sec:experiments}: the image deduplication procedure, the empirical calibration of the hazard-clustering gap tolerance $g$, and the calibration of the soft-constraint cost weight $w_{\text{cost}}$.

\subsubsection{Image Deduplication}
\label{sec:appendix-dedup}

Within the Crack, Patch, and Pothole categories, a base filename (chainage value) appearing with both a \texttt{.jpg} and a \texttt{.png} extension was confirmed, via MD5 hashing of the file contents, to be a byte-identical duplicate in every observed case; the \texttt{.jpg} copy is retained and the \texttt{.png} copy is dropped. The Background category exhibits no overlapping base names between its \texttt{.jpg} and \texttt{.png} files, so both are retained as distinct unit images in that category. This deduplication logic is shared with a related, out-of-scope state-transition data pipeline for Markov deterioration hazard modeling that processes the same raw PaveDistress files~\citep{Yasuno2026Scaling}, ensuring the two pipelines agree on what constitutes one unique unit image even though they differ in which categories they include (Section~\ref{sec:experiments}) and in what they compute from them.

\subsubsection{Gap-Tolerance Calibration}
\label{sec:appendix-gap}

The hazard-clustering gap tolerance $g$ (Section~\ref{sec:clustering}) is calibrated from the empirical distribution of consecutive-chainage gaps across all 4,922 deduplicated unit images (4,859 gaps between sorted, distinct chainage positions). Table~\ref{tab:gap-percentiles} reports selected percentiles. The median gap (2.07~m) corresponds closely to the nominal 2.1~m unit-image pitch (Section~\ref{sec:notation}), confirming near-continuous sampling within a surveyed run, while the long right tail (99th percentile $\approx$ 90.9~m) reflects occasional larger gaps within an otherwise continuous segment, as well as the much larger jumps between the four disjoint surveyed segments identified in Section~\ref{sec:sparsity} (which are excluded from this within-segment gap analysis by construction, since those jumps are orders of magnitude larger and are already handled separately by the data-run detection procedure). We selected $g = 11$~m because it is just above the 90th percentile (10.3~m) of the within-segment gap distribution, bridging the large majority of normal sampling gaps while remaining small enough not to merge genuinely separate distress locations connected only by a rare, unusually large gap.

\begin{table}[h]
\centering
\caption{Empirical percentiles of consecutive unit-image chainage gaps (metres), $n = 4{,}859$ gaps.}
\label{tab:gap-percentiles}
\begin{tabular}{lrrrrr}
\toprule
\textbf{Percentile} & \textbf{50th} & \textbf{75th} & \textbf{90th} & \textbf{95th} & \textbf{99th} \\
\midrule
Gap (m) & 2.07 & 4.13 & 10.33 & 18.59 & 90.87 \\
\bottomrule
\end{tabular}
\end{table}

\subsubsection{Cost-Weight Calibration}
\label{sec:appendix-cost}

The soft-constraint cost weight $w_{\text{cost}} = 0.12$ (Eq.~\ref{eq:benefit-cost}) was selected by inspecting the boundary behavior it induces at the minimum lot length $L_{\min} = 10.5$~m (Section~\ref{sec:minlot}). A cluster with a single crack observation ($R_j = 1$) padded to $L_{\min}$ has benefit $\mathrm{Benefit}_j = 1$ and cost $\mathrm{Cost}_j = 0.12 \times 10.5 = 1.26 > 1$, and is therefore excluded (Eq.~\ref{eq:decision}): a single, isolated crack observation does not by itself justify a repair-crew mobilization. A cluster with two or more crack observations within the same minimum-length lot, by contrast, has benefit $\mathrm{Benefit}_j \geq 2 > 1.26$, and is included. This operationalizes the qualitative design goal (Section~\ref{sec:introduction}) that crack repair should be pursued ``as much as cost-justified,'' treating a single weak signal as insufficient evidence while treating a repeated or spatially extended signal as sufficient. Of the 82 soft clusters observed in this case study, 37 (45.1\%) are single-crack-dominated clusters below this threshold and are excluded; the remaining 45 (54.9\%) satisfy it. As noted in Section~\ref{sec:limitations}, this threshold is a heuristic calibration rather than an economically estimated unit repair cost, and the skyline frontier construction (Section~\ref{sec:skyline-def}) is provided precisely so that a different threshold's consequences can be read off the same frontier without re-running the upstream clustering and padding steps.

\subsubsection{Computational Complexity}

Hazard clustering (Section~\ref{sec:clustering}) is a single linear pass over chainage-sorted trigger observations, $O(n \log n)$ including the sort. Minimum-lot-length expansion and patch-history attachment are each $O(J)$ and $O(J \log n)$ respectively (the latter using binary search against the sorted patch chainage array, Eq.~\ref{eq:patch-recurrence}). The hard/soft decision rule is $O(J)$. Lot merging is a single linear pass over chainage-sorted included lots, $O(J \log J)$ including the sort. Skyline frontier construction sorts the soft clusters by efficiency, $O(J \log J)$. The overall procedure is therefore $O(n \log n)$ in the number of unit images, well within the scale of the 4,922-image case study reported in this paper (wall-clock time on a single consumer workstation: under one second, excluding figure rendering).

\section{Supplementary: Geospatial Reinterpretation of Markov Deterioration Hazard Modeling in a Companion Study}
\label{sec:appendix-spatial-hazard}

This appendix elaborates, for completeness, on a point raised in Section~\ref{sec:hazard-reality} and Section~\ref{sec:limitations}: the companion Markov deterioration-hazard study~\citep{Yasuno2026Scaling} applies its state-transition hazard model to the same PaveDistress collection~\citep{PaveDistress2024} used here, but governs that model along chainage (spatial displacement) rather than along elapsed inspection time. We summarize this companion construction and its principal pitfall and findings below, since they bear directly on how a density-over-distance hazard signal can be derived from a single-epoch survey without requiring repeated inspections of the same physical location---precisely the constraint this paper's own method (Section~\ref{sec:introduction}) is designed to satisfy. Section~\ref{sec:appendix-scaling-equivalence} additionally presents the companion study's scaling-method comparison that evaluates, and underlies, this paper's own Definition 3 (Section~\ref{sec:notation}).

\subsection{Pseudo-Spatial State Transitions in Place of Temporal Inspection Pairs}
\label{sec:appendix-companion-setup}

The companion study's Markov hazard model requires paired ``before'' and ``after'' observations of the same location, separated by an elapsed interval, in order to estimate a continuous-time hazard rate. The ten bridge-inspection member cases that motivate that study supply this pairing naturally, from repeated inspections of the same structural member years apart. The PaveDistress collection, by contrast, is a single-epoch survey: each chainage position is photographed once, so no genuine repeated-inspection pair exists. The companion study therefore substitutes a \emph{pseudo}-temporal pair constructed from spatial adjacency: unit images are ordered by chainage position, grouped into 20-image pseudo segments, and each consecutive image pair within a segment is treated as a synthetic ``before/after'' transition with an assigned pseudo interval of $1.0 \pm 0.2$ years.

Formally, reusing this paper's own chainage-ordered unit-image index $i$ (Section~\ref{sec:notation}), let $S^{(i)} \in \{1, 2\}$ denote the companion study's binary health state of unit image $i$ (healthy, damaged). For each chainage-consecutive pair $(i, i+1)$ within a 20-image pseudo segment, the companion study defines a transition observation
\begin{equation}
\big(\mathrm{St1}^{(i)},\ \mathrm{St2}^{(i)},\ \Delta t^{(i)}\big) \;:=\; \big(S^{(i)},\ S^{(i+1)},\ \delta^{(i)}\big), \qquad \delta^{(i)} \sim \mathcal{U}(0.8,\,1.2)\ \text{years},
\label{eq:companion-pseudo-pair}
\end{equation}
where $\delta^{(i)}$ is a synthetic pseudo-interval assigned independently of the true chainage spacing $c^{(i+1)} - c^{(i)} = p = 2.1$~m (Section~\ref{sec:notation}). The index $i$ in Eq.~\ref{eq:companion-pseudo-pair} is a chainage-ordered \emph{position}, not a calendar time: consecutive transitions $(i, i{+}1)$ and $(i{+}1, i{+}2)$ correspond to three distinct physical locations $c^{(i)} \neq c^{(i+1)} \neq c^{(i+2)}$, one unit-image pitch apart, rather than two repeated observations of a single location at two different real times. Consequently, although the companion study's continuous-time hazard rate $\lambda$ is nominally parameterized as a function of elapsed calendar time $t$ since a location was last observed healthy, Eq.~\ref{eq:companion-pseudo-pair} shows that what actually varies across observations is the chainage index $i$ (equivalently, the displacement $c^{(i+1)} - c^{(i)}$), while $\Delta t^{(i)}$ is held to an approximately constant synthetic label. The hazard model is thus estimated not from the passage of real time at a fixed location, but from the geospatial density of damaged versus healthy observations encountered while moving along chainage---the same governing axis (chainage, not time) that underlies this paper's own repair-lot formulation (Section~\ref{sec:notation}). This is the sense in which the companion study's use of the Markov deterioration-hazard formalism is novel: it repurposes a transition-hazard estimator, ordinarily driven by calendar time between repeat inspections, to instead characterize how hazard density accumulates with distance traveled in a single pass, consistent with the chainage-centric reasoning already motivated on physical and logistical grounds in Section~\ref{sec:hazard-reality}. As in that section, this is explicitly a geospatial density characterization rather than a forecast of future deterioration at any specific location (Section~\ref{sec:limitations}).

\subsection{Binary Health Rank and a Structural-Leakage Pitfall}
\label{sec:appendix-leakage}

An initial version of the companion pipeline encoded each PaveDistress image's health state on a four-level ordinal rank, Background $<$ Crack $<$ Patch $<$ Pothole, carried over unchanged from the bridge-inspection member cases. This implicitly assumes that pothole formation passes through a crack stage and then a patch stage, i.e., a single ordered severity ladder. Because Pothole and Crack are independent distress categories with no such stepwise dependency---the same observation that motivates this paper's own asymmetric hard/soft treatment of the two categories (Section~\ref{sec:hazard-reality})---the companion study revised this to a binary rank (Background/Other/Patch = healthy; Crack/Pothole = damaged) and moved the damage-category information out of the rank and into two strictly-prior-history covariates, \texttt{patch\_history\_dummy} and \texttt{crack\_history\_dummy}, each equal to 1 if a Patch (respectively Crack) image occurs earlier in the same pseudo segment than the ``before'' observation of the transition being modeled, and 0 otherwise.

An initial implementation of the second covariate instead defined a \texttt{damage\_type\_dummy} directly from the \emph{outcome} (``after'') image's category, i.e., it encoded whether the transition's own result was a Crack. Because this covariate is a deterministic function of the quantity the hazard model is estimating the rate of, it produced severe quasi-complete separation: an estimated coefficient of $+14.6$ with non-finite (\texttt{NaN}) standard errors, and a degenerate expected-degradation curve of approximately 16{,}000 pseudo-years. Redefining the covariate as \texttt{crack\_history\_dummy}, computed only from information strictly prior to the ``before'' observation (matching the construction already used for \texttt{patch\_history\_dummy}), removed this leakage. The corrected model yields a sane coefficient magnitude and an average hazard rate of approximately 0.70 per pseudo-year, corresponding to an expected healthy-to-damaged transition time of about 1.4 pseudo-years. We record this pitfall here because it is a general risk when constructing any hazard-model covariate from a single-epoch, chainage-ordered image sequence: a covariate must be restricted to information available strictly before the position being predicted, or it risks silently encoding the label itself.

\subsection{Evaluating Definition 3: Scaling-Method Evidence from the Companion Study}
\label{sec:appendix-scaling-equivalence}

Definition 3 (Section~\ref{sec:notation}, Eq.~\ref{eq:zscore-hazard}) fixes the hazard-score scaling to z-score and states only that this choice is ``consistent with scaling-robustness evidence reported in a related, out-of-scope deterioration-modeling study,'' without this paper itself re-running a scaling comparison; the WCSP evaluated in Sections~\ref{sec:experiments}--\ref{sec:results} does not depend on this choice of scaling at all, since it operates on raw per-cluster counts, not on the z-score hazard map used only for visualization. This subsection supplies the evidence underlying that statement. The companion study compares four covariate-scaling transformations---max scaling, min-max scaling, z-score scaling, and a Box-Cox transformation with a training-derived positivity shift---across eleven member cases: ten bridge-inspection cases and this PaveDistress pavement case. Table~\ref{tab:appendix-scaling-bridgecases} reproduces the companion study's ten-bridge-case comparison, the primary basis for fixing z-score in Definition 3: z-score is the only method that converges for every one of the ten independent member cases, while simultaneously attaining the highest mean golden-test accuracy and the fastest median optimizer convergence time among the four methods. Definition 3's scaling choice is therefore non-arbitrary despite not being re-derived within this paper: it is fixed in advance from evidence of convergence reliability across a materially different deterioration-modeling task on the same general class of inspection data, rather than tuned to this paper's own case study.

\begin{table}[h]
\centering
\caption{Companion-study scaling-method comparison across the ten bridge-inspection member cases (underlying Definition 3's fixed z-score choice).}
\label{tab:appendix-scaling-bridgecases}
\begin{tabular}{lrrrr}
\toprule
\textbf{Scaling method} & \textbf{Converged} & \textbf{Pareto-sel.} & \textbf{Mean acc.} & \textbf{Median time} \\
\midrule
Max scaling & 9 / 10 & 0 & 0.5174 & 0.928 s \\
Min-max scaling & 9 / 10 & 1 & 0.5174 & 0.988 s \\
Z-score scaling & 10 / 10 & 7 & 0.5255 & 0.358 s \\
Box-Cox transformation & 4 / 10 & 3 & 0.4546 & 0.110 s \\
\bottomrule
\end{tabular}
\end{table}

Table~\ref{tab:appendix-scaling-pavedistress} reports the companion study's four-method comparison specifically for the binary-covariate PaveDistress case, i.e., the same case study used in this paper. All four methods converge here and are numerically indistinguishable on accuracy and log loss, because with two binary (0/1) covariates, max, min-max, z-score, and Box-Cox scaling are all affine-equivalent reparameterizations of the same columns; only optimizer wall-clock time differs. This case therefore contributes no independent evidence toward or against z-score specifically---it is scaling-invariant by construction---but it is consistent with, and not in tension with, the bridge-case evidence in Table~\ref{tab:appendix-scaling-bridgecases} that is the actual basis for Definition 3. We emphasize that this is a separate use of z-score scaling from the fixed scaling this paper itself applies to kilopost-level pothole/crack/patch counts for hazard-density visualization (Eq.~\ref{eq:zscore-hazard}): the companion study's z-score transform is fit to covariates of a logistic hazard regression, while Definition 3's z-score transform is fit to per-kilopost hazard counts for a descriptive map; the two uses share a scaling choice and the reliability evidence above, but operate on different quantities for different purposes.

\begin{table}[h]
\centering
\caption{Companion-study scaling-method comparison for the PaveDistress pavement case (binary covariates \texttt{patch\_history\_dummy}, \texttt{crack\_history\_dummy}; $n=195$ training rows, seed \texttt{20260924}).}
\label{tab:appendix-scaling-pavedistress}
\begin{tabular}{lrrr}
\toprule
\textbf{Scaling method} & \textbf{Converged} & \textbf{Accuracy} & \textbf{Log loss} \\
\midrule
Max scaling & Yes & 0.9876 & 0.0700 \\
Min-max scaling & Yes & 0.9876 & 0.0700 \\
Z-score scaling & Yes & 0.9876 & 0.0700 \\
Box-Cox transformation & Yes & 0.9876 & 0.0700 \\
\bottomrule
\end{tabular}
\end{table}

\end{document}